\documentclass[10pt,journal,compsoc]{IEEEtran}

\usepackage[english]{babel}
\usepackage{blindtext}
\usepackage[acronym]{glossaries}
\usepackage{acronym}
\usepackage{ltablex, enumitem, makecell, float, textcomp}
\usepackage{colortbl}
\usepackage{bibunits}
\usepackage{comment}
\usepackage{pifont}
\usepackage{amsmath}
\usepackage{tabularx}
\usepackage{graphicx}
\usepackage{subcaption}
\usepackage{hyperref}
\usepackage{amssymb}
\usepackage[table,xcdraw]{xcolor}
\definecolor{diff}{rgb}{0.0, 0.0, 0.0}
\definecolor{diffy}{rgb}{0.0, 0.0, 0.0}
\definecolor{dif}{rgb}{0.0, 0.0, 0.0}
\definecolor{irb}{rgb}{0.0, 0.0, 0.0}
\definecolor{cr}{rgb}{0.0, 0.0, 0.0}

\newcommand{\diff}[1]{\color{diff}{#1}~\color{black}}
\newcommand{\diffy}[1]{\color{diffy}{#1}~\color{black}}
\newcommand{\dif}[1]{\color{dif}{#1}~\color{black}}

\newcommand{\cre}[1]{\color{cr}{#1}~\color{black}}

\usepackage{cite}
\usepackage{balance}
\usepackage{booktabs}

\usepackage{multirow}
\definecolor{Gray}{gray}{0.85}
\definecolor{LightCyan}{rgb}{0.88,1,1}

\usepackage[ruled, lined, linesnumbered, commentsnumbered, longend]{algorithm2e}
\DontPrintSemicolon
\SetKwData{Result}{result}

\usepackage[utf8]{inputenc}
\usepackage{bbm}

\usepackage{array}
\newcolumntype{L}[1]{>{\raggedright\let\newline\\\arraybackslash\hspace{0pt}}m{#1}}
\newcolumntype{C}[1]{>{\centering\let\newline\\\arraybackslash\hspace{0pt}}m{#1}}
\newcolumntype{R}[1]{>{\raggedleft\let\newline\\\arraybackslash\hspace{0pt}}m{#1}}

\newcommand*\rot{\rotatebox{45}}

\usepackage{xspace}
\makeatletter
\DeclareRobustCommand\onedot{\futurelet\@let@token\@onedot}
\def\@onedot{\ifx\@let@token.\else.\null\fi\xspace}
\def\etc{\emph{etc}\onedot} 
\newcommand{\etal}{\textit{et al}. }
\newcommand{\ie}{\textit{i}.\textit{e}., }
\newcommand{\eg}{\textit{e}.\textit{g}., }
\acrodef{gnn}[GNN]{graphical neural network}

\acrodef{ml}[ML]{machine learning}
\acrodef{sota}[SOTA]{state-of-the-art}

\acrodef{fg}[FG]{Conference on Automatic Face and Gesture Recognition}
\acrodef{hog}[HOG]{histogram of gradients}
\acrodef{ss}[SS]{sister-sister}
\acrodef{bb}[BB]{brother-brother}
\acrodef{sibs}[SIBS]{brother-sister}

\acrodef{fs}[FS]{father-son}
\acrodef{ms}[MS]{mother-son}
\acrodef{fd}[FD]{father-daughter}
\acrodef{md}[MD]{mother-daughter}

\acrodef{gfgs}[GFS]{grandfather-grandson}
\acrodef{gmgs}[GMS]{grandmother-grandson}
\acrodef{gfgd}[GFD]{grandfather-granddaughter}
\acrodef{gmgd}[GMD]{grandmother-granddaughter}

\acrodef{ggfgs}[GGFS]{great \acs{gfgs}}
\acrodef{ggmgs}[GGMS]{great \acs{gfgd}}
\acrodef{ggfgd}[GGFD]{great \acs{gmgs}}
\acrodef{ggmgd}[GGMD]{great \acs{gmgd}}

\acrodef{fmd}[FMD]{father-mother-daughter}
\acrodef{fms}[FMS]{father-mother-son}

\acrodef{sdm}[SDM]{signal detection model}
\acrodef{roc}[ROC]{receiver operating characteristic}
\acrodef{nmse}[NMSE]{Normalized Mean Square Error}
\acrodef{det}[DET]{Detection Error Trade-off}
\acrodef{tp}[TP]{true-positive}
\acrodef{tn}[TN]{true-negative}
\acrodef{ap}[AP]{average precision}

\acrodef{map}[mAP]{mean AP}
\acrodef{cmc}[CMC]{Cumulative Matching Characteristic}
\acrodef{tpir}[TPIR]{true-positive identification rate}
\acrodef{frir}[FRIR]{false-reject identification rate}

\acrodef{tar}[TAR]{True Acceptance Rate}
\acrodef{far}[FAR]{False Acceptance Rate}
\acrodef{eer}[EER]{Equal Error Rate}

\acrodef{fn}[FN]{false-negative}
\acrodef{frr}[FRR]{false-reject rate}
\acrodef{fnr}[FNR]{false-negative rate}
\acrodef{fp}[FP]{false-positive}
\acrodef{fpr}[FPR]{false-positive rate}
\acrodef{tpr}[TPR]{true-positive rate}

\acrodef{fiw}[FIW]{\textit{Families In the Wild}}
\acrodef{mm}[MM]{{multimedia}}

\acrodef{fiwmm}[FIW MM]{\textit{FIW in Multimedia}}

\acrodef{tsk}[TSKIN]{\textit{Tri-Subject Kinship}}

\acrodef{kfw}[KinFaceW]{\textit{Kin-Faces in the Wild}}
\acrodef{kfvw}[KFVW]{\textit{KinFaceW Videos}}

\acrodef{rfiw}[RFIW]{\textit{Recognizing Families In the Wild}}

\acrodef{cnn}[CNN]{Convolutional Neural Network}
\acrodef{lut}[LUT]{Look-Up-Table}
\acrodef{fr}[FR]{face recognition}
\acrodef{gan}[GAN]{generative adversarial network}

\acrodef{vid}[VID]{Video ID}
\acrodef{svm}[SVM]{Support Vector Machine}
\acrodef{mid}[MID]{Member ID}
\acrodef{fid}[FID]{Family ID}
\acrodef{pid}[PID]{Photo ID}
\acrodef{roc}[ROC]{receiver operating characteristic}
\acrodef{nrml}[NRML]{Neighborhood Repulsed Metric Learning}

\acrodef{ge2e}[GE2E]{generalized end-to-end}

\acrodef{hog}[HOG]{histogram of oriented gradients}

\acrodef{nist}[NIST]{National Institute of Standards and Technology}

\acrodef{ta}[TA]{template adaptation}

\begin{document}
\title{Families In Wild Multimedia: A Multimodal Database for Recognizing Kinship}

\author{Joseph P. Robinson, Zaid Khan, Yu Yin, Ming Shao, Yun Fu

\IEEEcompsocitemizethanks{\IEEEcompsocthanksitem J.P. Robinson, Y. Yin, Z. Khan, and Y. Fu was with the Department
of Electrical and Computer Engineering, Northeastern University, Boston,
MA, 02115. E-mail: see http://www.jrobsvision.com\protect\\

\IEEEcompsocthanksitem M. Shao was with University of Massachusetts, Dartmouth, MA, 02747.}
\thanks{Manuscript received October 1, 2020; revised August 2, 2021.}}

\IEEEtitleabstractindextext{%
\begin{abstract}
\cre{Kinship, a soft biometric detectable in media, is fundamental for a myriad of use-cases.} Despite the difficulty of detecting kinship, annual data challenges using \dif{still-images} have consistently \dif{improved performances} and attracted new researchers. Now, systems \dif{reach} performance levels unforeseeable a decade ago, closing in on performances acceptable to deploy in practice. Like other biometric tasks, we expect systems can receive help from other modalities. We hypothesize that adding modalities to \ac{fiw}, which has only \dif{still-images,} will improve performance. Thus, to narrow the gap between research and reality and enhance the power of kinship recognition systems, we extend \ac{fiw} with multimedia (MM) data (\ie video, audio, and text captions). Specifically, we introduce the first publicly available multi-task MM kinship dataset. To build \ac{fiwmm}, we developed machinery to \dif{automatically collect, annotate, and prepare the data, requiring minimal human input and no financial cost.} The proposed MM corpus allows the \dif{problem statements} to be more realistic template-based protocols. We show significant improvements in all benchmarks with the added modalities. The results highlight edge cases to inspire future research with different areas of improvement. \ac{fiwmm} supplies the data needed to increase the potential of automated systems to detect kinship in MM. It also allows experts from diverse fields to collaborate in novel ways.
\end{abstract}

\begin{IEEEkeywords}
Kinship Verification, \dif{Face Recognition, Talking Faces, Visual Information, Audio, Multimodal,} Feature Fusion, Deep Learning, Template Adaptation, \dif{Biometrics, Multi-task, Support Vector Machines, Large-scale, Dataset, Convolutional Neural Network.}
\end{IEEEkeywords}}

\maketitle
\IEEEdisplaynontitleabstractindextext
\IEEEpeerreviewmaketitle

\acresetall
\acresetall 
\IEEEraisesectionheading{\section{Introduction}\label{sec:introduction}}
\IEEEPARstart{A}{utomated} kinship recognition assumes genetic relatedness between individuals is detectable by facial cues. \dif{The motivation behind the progress in this arduous task was} kinship datasets and advances in \ac{fr}~\cite{liu2017sphereface, masi2018deep}. The seminal work in visual kinship recognition introduced the first image dataset in 2010~\cite{fang2010towards}. \dif{Later came} \dif{more prominent} and challenging datasets, such as \ac{fiw}~\cite{robinson2018recognize} and \ac{tsk}~\cite{qin2015tri}. \dif{After that,} researchers proposed methods to match the level of difficulty in these kinship datasets~\cite{robinsonKinsurvey2020, robinson2020recognizing}.

\diff{Conventional \ac{fr} systems get, store, and recognize human faces automatically. Nowadays,} \ac{fr}, and the vast sub-problems that model visual knowledge from faces, have grown popular in speaker-based problems via audiovisual data (\eg speaker separation~\cite{ephrat2018looking}, speaker identification~\cite{nagrani2017voxceleb, chung2018voxceleb2}, cross-modal \dif{audio-to-visual} or vice-versa~\cite{Nagrani18a}, emotion recognition in \ac{mm}~\cite{Albanie18a, hao2020visual}, and more~\cite{Wiles18, Wiles18a}). The sudden surge of attention to audiovisual data has brought together specialists in biometrics to combine knowledge and find solutions that fuse multi-domain knowledge for best decision-making~\cite{song2019review, petridis2017end}. \dif{The benefits of added biometrics signals reaped in \ac{fr} intuitively can also enhance existing \ac{sota} in kinship recognition.}

\ac{fiw}~\cite{robinson2016families, robinson2018recognize}, the \textbf{largest} and \textbf{most comprehensive} image dataset for kinship recognition, includes 1,000 families with relationships labeled as tree structures: families have family and profile photos (\ie $\geq$ 12 photos), multiple members (\ie $\geq$ four members), and many faces (\ie 20 faces on average). The metadata includes first and last names, genders, bounding box coordinates of faces in photos, and the relationships with all other family members. \diffy{We chose 200 families of \ac{fiw} with $\geq$ two members in video data available online. The added biometrics in \ac{mm} (\eg the visual dynamics in videos, speech in audio, and spoken words in text captions) should \dif{complement} the facial images.}

\begin{figure*}
\centering
    \includegraphics[width=\linewidth]{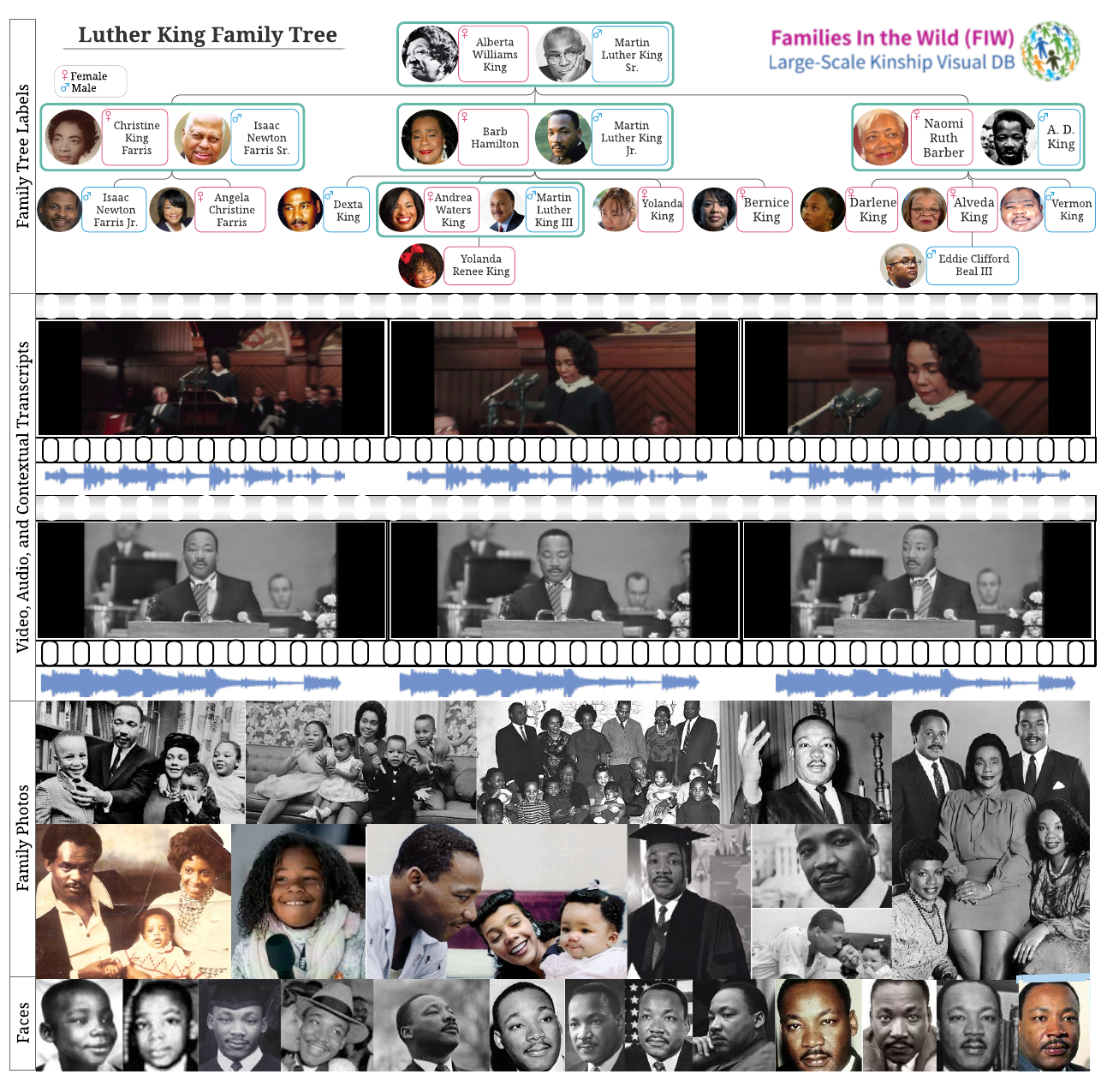}
  \caption{\textbf{Sample family of \ac{fiwmm}}. Top-to-bottom: \emph{family-tree labels} show faces of members in the immediate family, with subjects of the same generations in the same row; \emph{videos, audio, and contextual} exemplify sample video pairs of Dr. King Jr. and his daughter Andrea with tracklets of faces in the visual domain and audio data aligned frame-by-frame; \dif{randomly selected} \emph{family photos} that contain Dr. Luther King Jr. (note, cropped to fit); \emph{faces} of Dr. King Jr. from adolescence-to-adulthood. Multiple faces are available for most subjects. Best viewed electronically.}
  \label{fig:teaser}
\end{figure*}

\diff{The proposed shows that \ac{mm} improves the \ac{sota} in kinship recognition.} Our contributions to the \ac{fr}, biometric, anthropology, and \ac{mm} communities are three-fold.

\begin{itemize}
    \item {\verb|Built MM database|}: \diff{extended \ac{fiw} with MM data (\ie video tracks, speech segments, and text transcripts) via an automatic labeling scheme -- $>$600 audiovisual samples for $\geq$ two members of 200 \ac{fiw} families).} \dif{We reorganized \ac{fiwmm} for the added metadata and paired data at the subject and instance levels, respectively.}

    \item {\verb|Created protocols and benchmarks|}: a new paradigm for kinship recognition using \ac{mm} data. Specifically, the updated experiments to template-based, as there \dif{is} often \dif{a} variable number of samples per subject in real-world settings. We are the first to do kinship recognition \diff{with multimodal template-based data with family tree labels.} 

    \item {\verb|Proved the advantage of MM|}: we show an increase in system performance from \dif{still-images} to still-images and videos, and then, again, with speech signals added -- a clear benefit of each added modality is shown. Our analysis highlights the shortcomings of the different media types for future work to address.
\end{itemize}

This work will attract more scholars to kin-based problems using \ac{mm}. \ac{fiwmm} will be accessible online.\footnote{\href{https://web.northeastern.edu/smilelab/fiw/download.html}{https://web.northeastern.edu/smilelab/fiw/download.html}}

\begin{table}[t!]
\caption{Commonly used acronyms and variables.}
\label{tab:definitions}
\begin{tabular}{rll}
                               & FID    & Family ID                          \\
                               & MID    & Member ID                          \\
                               & PID    & Picture ID                         \\
\multirow{-4}{*}{\textbf{DB Terms}}     & VID    & Video ID                           \\
\rowcolor[HTML]{EFEFEF} 
\cellcolor[HTML]{EFEFEF}       & \acs{bb}     & \acl{bb}                    \\
\rowcolor[HTML]{EFEFEF} 
\cellcolor[HTML]{EFEFEF}       & \acs{ss}     & \acl{ss}                     \\
\rowcolor[HTML]{EFEFEF} 
\cellcolor[HTML]{EFEFEF}       & \acs{sibs}     & \acl{sibs}                     \\
\rowcolor[HTML]{EFEFEF} 
\cellcolor[HTML]{EFEFEF}       & \acs{fd}     & \acl{fd}                    \\
\rowcolor[HTML]{EFEFEF} 
\cellcolor[HTML]{EFEFEF}       & \acs{fs}     & \acl{fs}                        \\
\rowcolor[HTML]{EFEFEF} 
\cellcolor[HTML]{EFEFEF}       & \acs{md}     & \acl{md}                    \\
\rowcolor[HTML]{EFEFEF} 
\cellcolor[HTML]{EFEFEF}       & \acs{ms}     & \acl{ms}                      \\
\rowcolor[HTML]{EFEFEF} 
\cellcolor[HTML]{EFEFEF}       & \acs{gfgd}   &  \acl{gfgd} \\
\rowcolor[HTML]{EFEFEF} 
\cellcolor[HTML]{EFEFEF}       & \acs{gfgs}    & \acl{gfgs}              \\
\rowcolor[HTML]{EFEFEF} 
\cellcolor[HTML]{EFEFEF}       & \acs{gmgd}    & \acl{gmgd}          \\
\rowcolor[HTML]{EFEFEF} 
\cellcolor[HTML]{EFEFEF}       & \acs{gmgs}    & \acl{gmgs}              \\
\rowcolor[HTML]{EFEFEF} 
\cellcolor[HTML]{EFEFEF}       & \acs{ggfgd}  & \acl{ggfgd} \\
\rowcolor[HTML]{EFEFEF} 
\cellcolor[HTML]{EFEFEF}       & \acs{ggfgs}  & \acl{ggmgd}         \\
\rowcolor[HTML]{EFEFEF} 
\cellcolor[HTML]{EFEFEF}       & \acs{ggfgd}  & \acl{ggmgd}    \\
\rowcolor[HTML]{EFEFEF} 
\multirow{-15}{*}{\cellcolor[HTML]{EFEFEF}\textbf{Pair-types}} & \acs{ggmgs}  & \acl{ggmgs}        \\
                               & FMD    & father/mother-daughter             \\
\multirow{-2}{*}{\textbf{Tri-Pairs}} & FMS    & father/mother-son                  \\
\rowcolor[HTML]{EFEFEF} 
\cellcolor[HTML]{EFEFEF}                              & CMC    & Cumulative matching characteristic \\
\rowcolor[HTML]{EFEFEF} 
\cellcolor[HTML]{EFEFEF}       & DET    & Detection error trade-off  \\
\rowcolor[HTML]{EFEFEF} 

\cellcolor[HTML]{EFEFEF}       & FAR    & False-acceptance rate              \\
\rowcolor[HTML]{EFEFEF} 
\cellcolor[HTML]{EFEFEF}       & ROC    & Receiver operating characteristic \\
\rowcolor[HTML]{EFEFEF} 
\multirow{-5}{*}{\cellcolor[HTML]{EFEFEF}\textbf{Metrics}}     & TAR    & True-acceptance rate                       \\
                              & FR    & Facial recognition \\
                              & SVM    & Support vector machine \\
\multirow{-3}{*}{\textbf{Task / solution}}     & TA    & Template Adaptation                    \\
\rowcolor[HTML]{EFEFEF}                          & $P$    & Probe                              \\
\rowcolor[HTML]{EFEFEF}                          & $G$    & Search gallery                     \\
\rowcolor[HTML]{EFEFEF}                            & $X$    & Template                           \\
\rowcolor[HTML]{EFEFEF}                         & $x$    & Media sample                       \\
\rowcolor[HTML]{EFEFEF}                            & $x^+$    & Encoded positive sample               \\
\rowcolor[HTML]{EFEFEF}                             & $x^-$  & Encoded negative sample            \\  
\rowcolor[HTML]{EFEFEF}                             & $z$    & Encoded media sample               \\
\rowcolor[HTML]{EFEFEF}                              & $N_+$  & No. positive templates             \\
\rowcolor[HTML]{EFEFEF} \multirow{-7}{*}{\cellcolor[HTML]{EFEFEF}\textbf{Experimental}} & $N_-$  & No. negative templates            
\end{tabular}
\end{table}

\section{Related Works}\label{sec:related:work}
Attempts to recognize kinship in media began with domesticated animals (\eg dogs~\cite{hepper1994long} and sheep~\cite{POINDRON200799,poindron2007maternal}), as many species recognize their kin through various signals (\eg touch, smell, visual, and acoustics). From this, we hypothesized that information in \ac{mm}, besides image-level facial features, can be used \dif{to detect kinship in humans better.} Hence, knowledge extracted from imagery lacks much information. Modeling more complex signals (\eg face dynamics and speech in videos) attribute inheritable traits (\eg expressions, mannerisms, and accents). Nonetheless, getting \ac{mm} is expensive. We propose a method to collect data and show recognition improvements with \ac{mm}.

We next review existing work in visual kinship recognition and then research advances \dif{audiovisual} data for \ac{fr}.

\subsection{Kinship recognition}
Computer vision researchers began using faces to recognize kinship about a decade ago, where Feng~\etal proposed to model the geometry, color, and low-level visual descriptors extracted from faces to discriminate between KIN and NON-KIN~\cite{fang2010towards}. Others then formulated the problem as various paradigms (\eg transfer subspace learning~\cite{Xia201144, xia2012understanding}, 3D face modeling~\cite{vijayan2011twins}, low-level feature descriptions~\cite{zhou2011kinship}, sparse encoding~\cite{fang2013kinship}, metric learning \cite{lu2014neighborhood}, tri-subject verification~\cite{qin2015tri}, adversarial learning~\cite{zhang2020advkin}, ensemble learning \cite{wang2020kinship}, video understanding~\cite{zhang2014talking, sun2018video, georgopoulos2020investigating}, and, most recently, video-audio understanding~\cite{wu2019audio}). A common factor is the attempt to improve discriminatory power for classifying face pairs as KIN or NON-KIN; another commonality was the limited sample size and, thus, unrealistic experimental settings.

Robinson~\etal introduced a large-scale image dataset to recognize families in still-imagery called \ac{fiw}~\cite{robinson2016families,robinson2018visual}. \ac{fiw} has 1,000 families with an average of 13 family photos, \dif{five} family members, and 26 faces. It came with benchmarks for 11 pairwise types, with the top performance of the baselines being a fine-tuned CNNs (\ie SphereFace~\cite{liu2017sphereface} and Center-loss~\cite{wen2016discriminative}). This was the beginning of big data in kin-based vision tasks-- deep learning could then be used to overcome observed failure cases~\cite{wang2017kinship, wu2018kinship}. Furthermore, new applications such as child appearance \dif{prediction~\cite{ghatas2020gankin, gao2019will},} and familial privacy protection~\cite{mingaaai2020} were \dif{made} recently.

Nowadays, \ac{fiw} continues to challenge researchers with various views of image-based tasks. A myriad of methods showed the ability of machinery to use still-images to find kinship in a pair or group of subjects. Nonetheless, only so much information can be extracted from still-images. The dynamics of faces in video data (\eg mannerisms expressed across frames) have other information, and audio and text transcripts (\ie contextual data describing the speech and other sounds) can widen the range of cues we model to discriminate between relatives and non-relatives. We propose the first large-scale multimedia dataset for kinship recognition. Specifically, we used the familial data of the \ac{fiw} image database to build upon the existing resource~\cite{robinson2016families,robinson2018visual}, using the still-images of \ac{fiw} and adding video, audio, audiovisual, and text data of subjects. Note that the difference between the video and audio compared to the \dif{audiovisual} is that the former two are single modality and the latter has multiple modalities-- \dif{audiovisual} clips \dif{have talking-face tracks} aligned with the speech signal. After its predecessor, \dif{the} database was dubbed \ac{fiwmm}. En route to bridging \dif{research and reality,} we follow the protocols of \ac{fiw}~\cite{robinson2020recognizing}, but now can be template-based (\ie per \ac{nist} in~\cite{maze2018iarpa}). \diff{Figure~\ref{fig:teaser} depicts a sample family with \ac{mm} for MLK and his daughter.}

\diffy{An annual data challenge that was based on \ac{fiw} intended to attract more attention by supplying more structure and incentives for researchers to work on kin-based visual recognition.} \diffy{Namely, the \ac{rfiw} series,} which has been held annually since 2017~\cite{robinson2017recognizing}, and with the latest in 2020~\cite{robinson2020face}. There have been many great attempts on the still-images as a result~\cite{KinNet, AdvNet}. Recent surveys~\cite{qin2019literature}, tutorials~\cite{robinson2018recognize}, and challenges~\cite{lu2014kinship, lu2015fg,wu2016kinship,robinson2020recognizing} elaborate on \ac{rfiw} and the various submissions.

\begin{figure*}[t!]
    \centering
    \includegraphics[width=\linewidth]{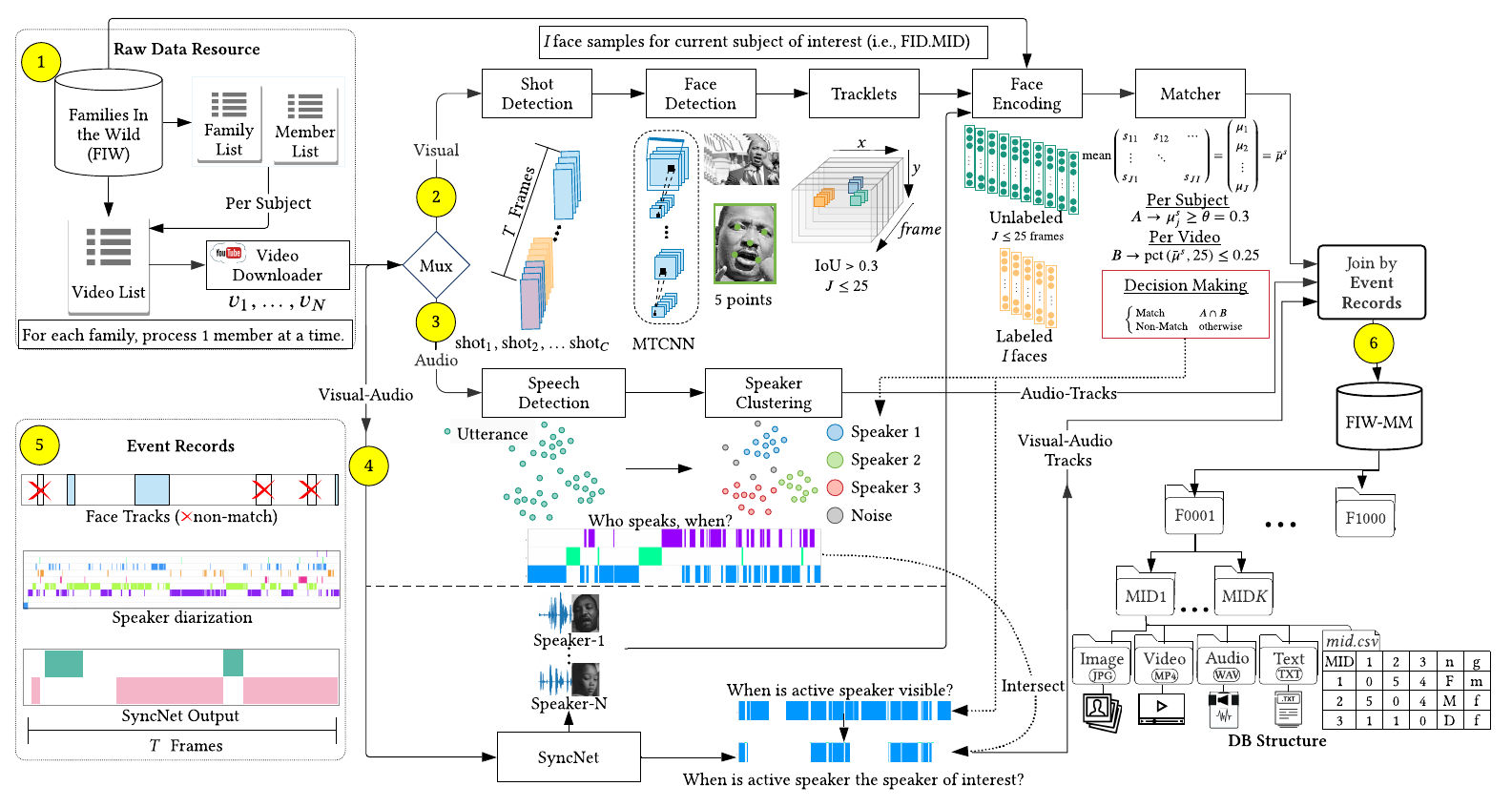}
    \caption{Our automated labeling framework. We extended a subset of families from \ac{fiw} with $N$ YouTube videos for $M$ subjects (\ie $M > 1$). Upon selecting the candidates, we generated and processed a list of videos per family, processing a single subject (\ie video-by-video) per iteration (1). First, we processed visual data: video shot detection (\ie C shots per video), detected face-tracks with $J$ samples per shot (\ie $J < 26$ faces): tracks continued for as many as 25 frames or if the IoU < 0.3 for neighboring bounding boxes. Next, ArcFace encoded all faces (\ie the $J$ unlabeled face-tracks and the $I$ labeled faces). Then, we compared labeled and unlabeled faces via cosine similarity, yielding a score-matrix ($J\times I$): fusing the score-matrix to a single decision (\ie \emph{Match} or \emph{Non-Match}) was done as follows: (i) the mean score of each frame across labeled samples produced a score vector of the size $J$; (ii) scores below the threshold (\ie 0.3) were marked \emph{Non-Match}; (iii) the 25-percentile of the samples marked as \emph{Match} represented the final, fused score; (iv) the fused score was compared to a threshold of 0.25 and, if greater than, the face-track was assumed to \emph{Match} the current subject of interest, which also represented in an event record to indicate frames with tracks that \emph{Match} (2). Returning to the \emph{Mux} and following the \emph{Audio} branch: we transformed audio signals to speech utterances, clustered speaker IDs into $K$ clusters for $K$ subjects, and recorded them as events (3). Returning to signal before \emph{Mux} (\ie before splitting the visual and audio modalities), we used the audiovisual data to detect frames with the active speaker visually displayed, which we also recorded as events (4). Next, the event records of size $T$ from steps 2-4 allowed us to produce a final record based on the agreement in records: we compared the event records to filter the samples of interest via logical AND (5). Finally, we stored the MM data in the \ac{fiwmm} database (6). Details for each step are in Section~\ref{subsec:datapipeline}. Best to view electronically.}\label{fig:workflow}
\end{figure*}

\subsection{Audiovisual data}
\dif{The big archetypal data} used for \dif{audiovisual} identification problems are Voxceleb~\cite{nagrani2017voxceleb} and Voxceleb2~\cite{chung2018voxceleb2}. Like \ac{fiwmm}, the datasets were acquired by extending image collections (\ie Voxceleb and Voxceleb2 extended of the VGGFace~\cite{Parkhi15} and VGGFace2~\cite{Cao18}, respectively). Currently, the primary usage of Voxceleb is in speaker-based tasks, such as using the \dif{audiovisual} data to detect and classify the speaker by the \emph{who} and the \emph{when}~\cite{ephrat2018looking}. Other \dif{speaker-centric} problems have been proposed using the Voxceleb collections, like to enhance speech signals~\cite{afouras2018conversation}, to detect \emph{when} and \emph{where} the speaking face is visible~\cite{Chung17a}, and when the audio and mouth motions infer the lips and sound are in sync~\cite{Afouras18b}. Nonetheless, the lip-reading task predates the larger Voxceleb with older lip-reading datasets~\cite{Chung16, Chung17}. 

It is worth highlighting that these \dif{audiovisual} databases were instrumental in applied research as well (\eg generating talking-faces~\cite{Chung17b}, where the input is a still-image face and a stream of audio, and the output \dif{frames} mocking the audio with the faces as if the input face was regurgitating the audio clip). In~\cite{wiles2018x2face}, face frames were generated from a still-image and audio clip, with pose information added as a control signal for the synthesized output. Furthermore, Voxceleb predicted emotion labels via its signals to automatically infer ground-truth~\cite{Albanie18}.

\dif{Minimal} attempts \dif{to} recognize kinship in \dif{audiovisual data have been made.} Most relevant was in~\cite{wu2019audio}, where the authors \dif{collected} 400 pairs. Wu~\etal certainly proved the core hypothesis of this work-- multimedia can enhance our ability to automatically detect kinship in humans-- as was clearly shown in their work~\cite{wu2019audio}. However, the sample size was limited in the number of pairs and the types of labels, as there is no family tree structure nor multiple samples per member (\ie age-varying), as is the case in our much more extensive and comprehensive \ac{fiwmm}.

\section{The FIW-MM Database}
\ac{fiwmm} \dif{extended the existing paired faces} of \ac{fiw} \dif{via} an automated labeling pipeline that \dif{allowed the proposed data} to be acquired with no financial cost and minimal human input (Figure~\ref{fig:workflow}). Specifically, \ac{fiw} allowed us to find \emph{who}, \emph{when}, and \emph{where} a family member appeared in the video. We chose 200 (\dif{of the} 1,000) \ac{fiw} families \dif{with} 2-5 members in 1-3 YouTube videos. Hence, \ac{fiwmm} \dif{consists of} 550 subjects in 660 videos. We obtained three types of data per subject per video: \dif{(1)} non-speaking face tracks (\dif{\ie}visual only), \dif{(2)} speech segments (\dif{\ie}audio only), and \dif{(3)} face tracks of \dif{the} speaker (\dif{\ie}\dif{audiovisual).} Timestamps \dif{were} set for the start and end frames, along with the bounding box information for the \dif{face tracks.} In this \dif{way,} overlap in samples \dif{was} identifiable.

Let us next cover the data specifications, along with a detailed description of our automatic labeling pipeline.

\diff{\subsection{Specifications}}
The goal was to extend \ac{fiw} in the number of samples and the types of media. Plus, improved experimental protocols. Recalling that \ac{fiw} supplies name metadata and face images for multiple members of 1,000 families~\cite{robinson2018recognize}, we used this from the 200 families. For quick reference, common acronyms and symbols are in Table~\ref{tab:definitions}.
\begin{table*}[!t]
    \centering
    \caption{Database statistics. Types are split based on the span in the generation of the relationship.}\label{tab:data-counts}
     \resizebox{\textwidth}{!}{%
    \begin{tabular}{r c c c c c c c c c c c c c c c c}
        & \multicolumn{3}{c}{\cellcolor{blue!15}\textbf{{1$^{st}$ Generation}}} & \multicolumn{4}{c}{\cellcolor{red!15}\textbf{{2$^{nd}$ Generation}}}  & \multicolumn{4}{c}{\cellcolor{green!15}\textbf{{3$^{rd}$ Generation}}}& \multicolumn{4}{c}{\cellcolor{purple!15}\textbf{{4$^{th}$ Generation}}}\\
        
         & \rot{\emph{\acs{bb}}} & \rot{\emph{\acs{ss}}} & \rot{\emph{\acs{sibs}}} & \rot{\emph{\acs{fd}}} & \rot{\emph{\acs{fs}}} & \rot{\emph{\acs{md}}} & \rot{\emph{\acs{ms}}} & \rot{\emph{\acs{gfgd}}} & \rot{\emph{\acs{gfgs}}} & \rot{\emph{\acs{gmgd}}} & \rot{\emph{\acs{gmgs}}} & \rot{\emph{\acs{ggfgd}}} & \rot{\emph{\acs{ggfgs}}} & \rot{\emph{\acs{ggmgd}}} & \rot{\emph{\acs{ggmgs}}} & {\emph{Total}}\\ 

       \textbf{\# Subjects}& 883 & 824 & 1,542 & 1,914 & 1,954 & 1,892 & 2,041 & 426&463 & 483 & 526 & 39&30 &45  &  37&13,099\tabularnewline
       
       \textbf{\# Families}&  345&  334&  472&  666& 676 & 665 &  670&154 & 174& 178 & 191 & 9&10 &11  &10  & 953 \tabularnewline

        \textbf{\# Still-images}&40,386 & 31,315 &46,188  & 83,157 & 89,157 & 57,494 &63,116 &8,007 & 6,775 & 6,373 &6,686 &408 & 410 &798& 797 &441,067  \tabularnewline

       \textbf{\# Clips}& 123 & 79 & 81 & 155 & 134 & 147 & 138 &16 & 18& 25 & 15 &2 &4 & 0 & 0 & 937\tabularnewline
       
       \textbf{\# Pairs}& 641 & 621 & 1,138  & 1,151 & 1,253 & 1,177  & 1,207  &263 &280 & 292 & 324  & 28 &18 & 36  &28  & 8,457
    \end{tabular}}
\end{table*}

Following \dif{the} convention of the original \ac{fiw}~\cite{robinson2016families}, the indices of the topmost level of the database were unique \acp{fid}, which have $M$ \acp{mid} for $M$ family members. Each \ac{fid} is represented by a relationship matrix of size $M$ (\ie a row and a column added per \acp{mid}), along with the gender information for each member. Thus, \ac{fiw} is organized with 1,000 folders (\ie one per \ac{fid}, F0001-F1000), each with a relationship matrix that relates the $M$ \acp{mid} that have face data stored in subfolders (\eg family F0009 \dif{has $M=7$,} meaning the folder contains 7 \ac{mid} folders, MID1, ..., MID7). The same scheme is adopted for \ac{fiwmm}, with the difference being that the \ac{mid} folders now have a folder per media type (\ie subdirectory \emph{images} added to \ac{mid} folder for which existing face images are now stored). Besides, folders for other media types were added.

\subsection{Data pipeline}\label{subsec:datapipeline}
Inspired by earlier work, such as \ac{fiw}~\cite{robinson2018visual} (\ie labeling families) and VoxCeleb~\cite{nagrani2017voxceleb} (\ie labeling \dif{audiovisual} data), we formed the basis for our pipeline. From this, three modality-specific branches processed the data independently. Specifically, the branches process the \emph{video}, \emph{audio}, and \dif{audiovisual} modalities. \diff{Furthermore, branch-specific events are recorded by the frame number and used to summarize instances that are propagated between records.}

The following subsections are numbered according to the yellow circle callouts in Figure~\ref{fig:workflow}.

\noindent\textbf{\diff{Step 1. }Raw data resource.}
A \dif{single-family} with at least two members on \dif{YouTube} was selected (\ie families with only one member with MM would be useless). Video URLs were queried per unique \acp{vid} (\ie $v_1, \dots, v_N$ for $N$ videos). The videos were either interview-style (\eg with only a news anchor in a plain room answering scripted questions) or face-time clips (\ie self-recordings with the subject speaking directly to the camera). Also, the ethnicity for subjects \dif{was} manually collected.

We used Pypi's youtube-dl\footnote{\href{https://github.com/ytdl-org/youtube-dl/}{https://github.com/ytdl-org/youtube-dl/}} to download YouTube videos by URL: each renamed to its \ac{vid} and stored as \dif{an} MKV file, along with TXT with captions when available. The raw data consists of the original MKV for the \emph{\dif{audiovisual}} branch, audio-only (WAV), and \dif{visual-only} (MP4) extracted with \dif{\emph{FFmpeg}} assuming 25 FPS.

\begin{table*}[t!h]

    \centering
    \caption{\textbf{Task-specific counts.} Per task (\ie verification in top row and identification bottom three), individuals (\textbf{I}), families (\textbf{F}), images (\textbf{S}), video-clips (\textbf{V}), audio snippets (\textbf{A}), audio snippets (\textbf{VA}) in the set of probes (\textbf{P}), gallery (\textbf{G}), and total (\textbf{T}).}
\begin{tabular}{m{.5mm}ccccc@{\hskip .1in}ccccc@{\hskip .1in}ccccc}
        &  \multicolumn{5}{c}{\cellcolor{blue!15}\textbf{Train}}{\hskip .1in}& \multicolumn{5}{c}{\cellcolor{red!15}\textbf{Val}}{\hskip .1in}  & \multicolumn{5}{c}{\cellcolor{green!15}\textbf{Test}}\tabularnewline
        
         &  \emph{I} & \emph{F} & \emph{S} & \emph{V} & \emph{A} & \emph{I} & \emph{F} & \emph{S} & \emph{V} & \emph{A} & \emph{I} & \emph{F} & \emph{S} & \emph{V} & \emph{A}\tabularnewline\tabularnewline[-.5em] 
  
\rowcolor{gray!0} \textbf{T} & 2,976 & 571 & 16,464 & 290 & 7,217 & 955 & 190 & 5,458 & 72 & 3,308 & 972 & 192 & 5,231 & 91 & 1,775
  \tabularnewline\tabularnewline[-.5em]

   \textbf{P} & 571 & 571 & 3,039 & 47 & 1,843 & 190 & 190 & 1,334 & 16 & 789 & 192 & 192 & 993 & 23 & 876       \tabularnewline
   \textbf{G} & 2,475 & 571 & 13,571 & 244 & 5,581 & 791 & 190 & 4,538 & 56 & 2,519 & 800 & 192 & 4,705 & 69 & 899  \tabularnewline
    \textbf{T} & 3,046 & 571 & 16,610 & 291 & 7,424 & 981 & 190 & 5,872 & 72 & 3,308 & 992 & 192 & 5,698 & 92 & 1,775
\end{tabular}\label{tbl:counts} 
\end{table*}

\noindent\textbf{\diff{Step 2.} Event records.} 
Before branching, blank (sequential) tabular records were instantiated for the duration of the video-- one record per \dif{branch} denoted 3-5 (\ie audio, visual, and \dif{audiovisual} event records). These are later compared to share knowledge between the branches to help filter out samples of the subject of interest. In essence, the mutual information across records at a given instance (\ie frame-step) \dif{is} used to imply matches, contradictions, and non-matches across modalities (\ie a means to propagate labels across modalities). The usage of set theory helps to both confirm \emph{matches} and filter out \emph{non-matches}. Although unique to our problem, the concept of using logic across events to parse videos has been done (\eg \cite{haq2019movie}); however, opposed to high-level semantics like types of objects, we care about the more straightforward tasks of \dif{a} face or no face, speech or not, visible or unseen, and then the same or different subject. Furthermore, doing this increases the random chance by adding more evidence \dif{across records.}

\diff{Specifically, and for one video simultaneously, the} \dif{three-event} records are instantiated with zero event entries. Then, events of each branch are recorded. The type of event \dif{is} later described for each branch: like branch 3, visual branch, the events are face tracks that is a \emph{match} with BB coordinates recorded for all frames of the shot; the audio branch consists of instances that each of the $k$ speakers \dif{pronounces} utterances; the record for the \dif{audiovisual} branch \dif{is} when the speaker is visible in the current frame. Then, by propagating the frame number that the subject of interest appears, the other audio instances with the same speaker and the frames where the speaker \dif{is} visible can be inferred by finding intersecting events between the three records.

\noindent\textbf{\diff{Step 3.} Visual branch.} 
A video was first parsed into scenes by using two global measures and with the assumption that, statistically, neighboring frames of the same shot will match as close as 90\% when comparing HSV (\ie color) and local binary patterns~\cite{ahonen2006face} (\ie texture) features. The features were extracted and used to parameterize two probabilistic representations per frame (\ie one per feature). Then, neighboring frames were measured via KL-Divergence and compared with a threshold of 0.1~\cite{sanchez2017multimedia}. A pair of frames that scored below the threshold were assumed to be shot boundary frames for $V$ videos of size $T\text{,}$ \ie \dif{$v_t\in\{1, 2, \dots, C\}$} represents all shots detected in the $i$-th video. The first, \dif{last,} and the frame closest to the centroid (\ie in color and texture) were used as the shot: the three frames were run through \dif{an} MTCNN face detector~\cite{zhang2016joint}, and for those without at least one face detection were assumed face-less scenes. Otherwise, the clips with face detections had as many as five more frames sampled and passed to the face detector, and then all faces were encoded and compared to the faces for members of the respective family in \ac{fiw}. Events were then recorded for the face tracks that \emph{matched} the subject of interest. Note that this could to quickly drop unwanted \dif{data} and reduce noise in events assumed by the other branches.

Faces were encoded with ArcFace via the architecture, settings, and \emph{matcher} of~\cite{deng2019arcface}. Specifically,

\begin{equation}\label{eg:matcher}
    d_{boolean}({x}_i, {x}_j) = d({x}_i, {x}_j) \leq \theta,
\end{equation}
where the \textit{matcher} $d_{boolean}$ compared the \dif{$i^{th}$-to-$j^{th}$} face encoding of \ac{fiw}~\cite{LFWTech}. Hence, $d_{boolean}$ is the decision boundary in score space-- if threshold $\theta$ is satisfied, assume \emph{match}; else, \emph{non-match}. Note that it is currently assumed that $i$ and $j$ are from different sets (\ie with $J$ labeled samples from \ac{fiw} and $I$ face detections from newly collected). The \emph{matcher} in Eq~\ref{eg:matcher} was set as cosine similarity the closeness of the L2 normalized~\cite{wang2017normface} \dif{features} by
\dif{$
d_{boolean}({x}_i, {x}_j) = 1 - d({x}_i, {x}_j) = \frac{z_i\cdot z_j}{||z_i||_2||z_j||_2} > \theta
$,} where $z$ represents media encoding. We set $\theta=0.2$ manually for a high recall. This process - including the usage of ArcFace to encode faces - is the \textit{matcher} used throughout.

In the end, scenes having the subject of interest had all its frames processed by the MTCNN-- the bounding box coordinates, fiducials (\ie 5 points), and confidence scores were recorded for each step. We then processed the BB coordinates to ensure continuity, dropping those without it: the ROI was set on the prior face location, and the IoU was calculated frame-by-frame, which had to surpass a threshold of 0.3. Finally, up to 25 (\ie $J$) faces were sampled per \dif{track} and passed to \dif{$d_{boolean}$} with each of the $I$ labeled faces (\ie producing $J\times I$ score matrix). The mean across $I$ samples produced a single score per the $J$ faces, at which point the value at the 25-percentile was compared to a higher threshold of $\theta=0.25$. Note, the fusion of scores was done to both consider all the existing labeled faces \dif{equally} while avoiding much weight on any of the few (of $J$) potentially low-quality faces. This step alone yielded many face tracks of type \emph{match} with \dif{high} confidence.

\noindent\textbf{\diff{Step 4.} Audio branch.}
Audio signals \dif{were} extracted from the videos and saved as high-quality WAV files. For the entire audio \dif{clip,} we did speaker diarization-- the process of detecting utterances of speech (\ie the \emph{when}), to then form $K$ clusters for $K$ speakers \textbf{(\ie the \emph{who})}. Note that the clusters are arbitrarily assigned IDs per video. In the end, a speech detector determined the \emph{when}, and clusters determined the number of speakers and, thus, the speech segments from the \emph{who}. The utterance detector used was SpeechRecognizer\footnote{\href{https://github.com/Uberi/speech_recognition}{https://github.com/Uberi/speech\_recognition}}, and clusters were based on models from~\cite{chung2020in}. 

\noindent\textbf{\diff{Step 5.} \dif{Audiovisual} branch.} 
\dif{The aim was} to detect when the speaker is in the field of view. Thus, the purpose was to find the frames for which the face and speech were aligned. An intuitive way would be to relate the faces detected, lip motions, and audio-- which is at the core of many speaker ID methods in \ac{mm}~\cite{zhu2020deep}. For this, videos were processed using SyncNet~\cite{Chung16a}), pre-trained from~\cite{li2017targeting}. The output \dif{was} the BB at frames for which the audio aligned to that face. 

\dif{The faces tracks belonging to the subject of interest were found and used to record events.} Furthermore, these events, compared with the audio events, allowed the cluster holding all speech utterances of the subject of interest formed at Step 4 to be figured out.

\noindent\textbf{\diff{Step 6.} Outputs and DB structure.} The face tracks, audio segments, and \dif{audiovisual} tracks of the subject of interest are output and retractable via the final event record (\ie the three records merged with only positive instances). 
Thus, \dif{the} overlap between audio and \dif{audiovisual} infers which cluster of audio segments belongs to the respective active speaker, while the processing of the visual and comparison to the original \ac{fiw} allowed for the subject of interest to match up with the active speaker.
Any overlap in the data found in the \emph{visual branch} or \emph{audio branch} versus the \emph{\dif{audiovisual}} \dif{was} removed as duplicated instances. The face tracks, speech segments, and clips with aligned \dif{audiovisual} were added to the folder named after media type in the \ac{mid} folders, along with the final event record (\ie the event record is enough to parse raw data). \diff{Then, the database is $N$ FID folders with $M$ \dif{MID folders} and a relationship matrix.}

\begin{table*}[h!t]
      \centering
        	\small
        	\caption{{\acf{tar} (\%) at specific \acf{far}. Scores are for template-based settings: \emph{Image} only (left column), \emph{Image} + \emph{Video} (middle), and \emph{Image} + \emph{Video} + \emph{Audio} (right). Higher is better.}}\label{subtab:task1:results}
\begin{center}
\begin{tabular}{R{14mm} m{2.3mm}m{2.3mm}m{3.3mm} m{2.3mm}m{2.3mm}m{3.3mm}  m{2.3mm}m{2.3mm}m{3.3mm} m{2.3mm}m{2.3mm}m{3.3mm} m{2.3mm}m{2.3mm}m{3.3mm} m{2.3mm}m{2.3mm}m{3.3mm} m{2.3mm}m{2.3mm}m{3.3mm} m{2.3mm}m{2.3mm}m{2.3mm} }
    \emph{\ac{far}} & 
    \multicolumn{3}{c}{\emph{\acs{bb}}{\cellcolor{red!0}}}& \multicolumn{3}{c}{\emph{\acs{ss}}}& \multicolumn{3}{c}{\emph{\acs{sibs}}}& 
    \multicolumn{3}{c}{\emph{\acs{fd}}}& \multicolumn{3}{c}{\emph{\acs{fs}}}& \multicolumn{3}{c}{\emph{\acs{md}}}& \multicolumn{3}{c}{\emph{\acs{ms}}} & 
    \multicolumn{3}{c}{Avg.}\tabularnewline
        0.5  &97.8&97.8&\textbf{98.2} &91.5&92.3&\textbf{92.7} &91.7&90.8&\textbf{91.5} &79.8&77.8&\textbf{79.9} &85.3&85.3&\textbf{87.1} &90.6&88.8&\textbf{91.4} &81.3&82.6&\textbf{85.2} &88.3&87.9&\textbf{89.8}\tabularnewline

        0.3  &94.1&94.1&\textbf{95.3} &88.0&87.2&\textbf{90.1} &82.9&83.9&\textbf{85.7} &63.5&66.5&\textbf{69.3} &77.1&79.1&\textbf{81.5} &82.4&82.3&\textbf{85.0} &68.9&70.1&\textbf{73.4} &79.6&80.4&\textbf{81.6}\tabularnewline
        
        0.1  &88.1&87.4&\textbf{88.4} &76.1&76.1&\textbf{79.1} &68.7&68.2&\textbf{70.2} &34.5&36.9&\textbf{42.9} &54.3&54.3&\textbf{58.2} &62.2&63.1&\textbf{69.4} &46.1&46.5&\textbf{50.1} &61.4&61.8&\textbf{64.9} \tabularnewline
        
        0.01  &70.4&70.4&\textbf{73.6} &54.7&55.6&\textbf{59.9} &44.2&46.1&\textbf{52.4} &5.9&7.9&\textbf{12.9} &23.6&24.0&\textbf{32.1}&28.3&31.3&\textbf{40.6} &11.6&13.3&\textbf{21.0} &34.1&35.5&\textbf{41.1} \tabularnewline
        
        0.001  &54.8&57.0&\textbf{61.1} &47.9&48.7&\textbf{52.4} &29.5&29.0&\textbf{33.7} &2.0&2.5&\textbf{7.7} &9.3&10.9&\textbf{14.1} &14.2&14.6&\textbf{18.5} &3.3&4.6&\textbf{7.8} &23.0&23.9&\textbf{30.1} \tabularnewline
\end{tabular}
\end{center}
\end{table*}

\section{Problem Statements}\label{sec:experimental}
Following the recent \ac{rfiw} data challenge~\cite{robinson2020recognizing}, we benchmark two kin-based tasks:  kinship verification and search \& retrieval of family \dif{members.} In \dif{\ac{rfiw}} and the contemporary works in kinship recognition, protocols are \dif{image-based,} \ie unimodal, \emph{single-shot} experiments. By contrast, \ac{fiwmm} supports \dif{multimodal,} with added samples and media types (Table~\ref{tab:data-counts}). Provided the \ac{mm}, protocols are template-based, like with IJB-A~\cite{maze2018iarpa}.

Kinship verification has been the primary focus \dif{of} experiments. More recently, the emergence of the more \dif{challenging} but more \dif{practical} task of \emph{searching for missing family members} was supported~\cite{robinson2020recognizing}. We benchmark \ac{fiwmm} for both tasks. With the difference being the template-based~\cite{maze2018iarpa}-- approaching settings of operational use-cases.

\dif{We next provide details for template-based recognition tasks. Then, we describe the tasks:} first kinship \dif{verification and} then search \& retrieval of family members. The paragraph structure \dif{stays consistent:} task overview, data splits and settings, and task-specific metrics.

\subsection{Definitions and protocols}
Template $X$ holds all media for a subject (\ie images, videos, \dif{audio clips).} Hence, $X$ consists of samples $x$, with a total of $N$ templates for the $n^{th}$ independent piece of media $x_n$, with a size of the sum of all data samples, $N=N_I+N_V+N_A$. \diffy{We stored features for all samples. Now, the templates include feature vectors, \ie $\mathcal{F}(x) =z$, where $\mathcal{F}$ maps face to the feature space via $\mathcal{F}(x)\in\mathbb{R}^\mathrm{d}$, and with $d$ being the size of the vector.} \dif{We set a subject-specific template as} probe $P$ (\ie query) or reference $Q$ (\ie hypothesis). For 1:$N$, known subjects were from a known family, \dif{and} templates enrolled in gallery $G$. Then, at inference, the goal was to match an unseen probe $P$ with $G$, where $|G|$ refers to \dif{the} number of templates enrolled \diffy{in the gallery.} \dif{As mentioned later,} for 1:1\diffy{-based evaluation,} we assume $|G|= 1$ (\ie compared the template for $P$ with the template of $G$ to decide whether the pair is \emph{KIN} or \emph{NON-KIN}). \diffy{For $|G|>1$,} the one-to-many 1:$N$ search \& retrieval task outputs ranked lists of template IDs sorted by the likelihood of being a blood relative (\eg if $|G|= 10$, then template of $P$ will compare with the \dif{ten} templates of $G$, to generate a list of indices [1, 10] ordered by score compared with probe \dif{$P$).}

As mentioned, \dif{we treat} an audio segment (\ie a clip of \dif{a} subject speaking without interruptions or \dif{significant} pauses) as a single piece of media $x\text{,}$ which is fused to a single representation by averaging across frames. Note that a video may consist of several disjoint \dif{tracks,} visual, audio, and \dif{audiovisual} (\ie aligned). Thus, there are many independent media samples for both the visual and audio modalities. Again, this is the set of media \dif{making up template $X$ per subject, which contains} various media samples $x$, such that the $j^{th}$ subject can be represented by $k$ media samples as follows: $X_j={\mathcal{F}_t(x_1), \mathcal{F}_t(x_2), \dots, \mathcal{F}_t(x_k)}$, where $t$ corresponds to the media type and, hence, the corresponding encoder. From this, $|X_j|$ is the total number of \dif{features} for subject $j$. The \emph{gallery} $G$ consists of a set of subjects by $G=\{(X_1, y_1)^l, (X_2, y_2)^l, \dots, (X_n, y_n)^l\}$, where $y$ are identity labels for each of the $N$ subjects, and $l \in \{1,2, \dots L \}$ are ground-truth for $L$ families. \dif{Each tuple also contains} a tag representing the set of $L$ families (\ie $(X_j, y_j)^l$), where $l\in\{1, 2, \dots, L\}$. Further partitioning of the data is done per \dif{the} requirements of a task. For instance, for the verification, the $i^{th}$ pair of tuples from the same family $\mathbb{P}_m=((X_i, y_i)\bigcap(X_j, y_j))$, where $i\neq j$, inherit labels KIN (\ie \emph{match}) and relationship type. 

Each task consists of a ($\approx$60\%) train, ($\approx$20\%) validation, and ($\approx$20\%) test set. These sets are disjoint in family and subject IDs-- sets were generated by splitting family labels, with partitioning that remained constant for all tasks.

\subsection{Kinship verification}\label{sec:track1}
Kinship verification is the \dif{simplest} of tasks in this challenging and complex \dif{problem space.} Hence, this one-to-one paradigm is the main view vision researchers have tackled. \diff{The task is to figure out whether a \dif{face pair} are blood relatives (\ie \emph{true kin} or \emph{match}). This face-based problem inherits all the challenges of conventional \dif{\ac{fr},} such as variations in lighting, pose, and occlusion (\eg sunglasses or beards). Additionally, several challenges specific to kin-based \ac{fr} are posed by age variations, face pairs that contradict directional relationships (\eg grandparent-grandchild, where the face image of the grandfather was from a younger age and that of the grandchild was older), bias \dif{across} subgroups like in \ac{fr}~\cite{robinson2020face} but amplified for the kin-based data.}

The most fundamental question asked in kinship \dif{verification} and re-asked in all other kinship discrimination \dif{tasks} is whether a face pair is related. Therefore, kinship verification is \dif{the} boolean classification of pairs (\ie ${y}\in\{\emph{KIN}\bigcup\emph{NON-KIN}\}$). Typically, knowledge of the relationship type is assumed. Thus, supplied the output of the model for a given pair is \emph{KIN}, then the specific type is implied. Future efforts could incorporate relationship-type signals to advance capabilities of kinship detection systems; however, and as stated upfront, verification provides the simplest of all the benchmarks and, up until now, is the most popular~\cite{robinson2020recognizing}.

\subsubsection{Data splits and settings}
Conventionally, a query consists of a single face image $x_1$ paired with a second face image $x_2$ (\ie a one-shot, boolean classification problem with labels $y\in\{\text{KIN}, \text{NON-KIN}\}$). Put formally, given a set of face-pairs $(x_1, x_2)_s^l$, where the number of sample pairs $s\in\{1, 2, \dots, S\}$ of relationship-type (\eg \emph{mother}-\emph{son}). A set of pair-lists $\mathbb{P}=\{[(x_1, x_2)_1^l]_\mathbbm{1}, [(x_1, x_2)_2^l]_\mathbbm{1}, \dots ,[(x_1, x_2)_S^l]_\mathbbm{1}\}$ for the $L$ types, and with the label determined by the indicator function $\mathbbm{1}$ for a single pair $\mathbb{P}_s\rightarrow\{ 0,1\}$, \ie

\begin{equation}\label{eq:indicator}
\mathbbm{1}(\mathbb{P}_s) = 
\begin{cases} 
            0 & \hspace{5mm} \emph{NON-KIN} \\
            1 & \hspace{5mm} \emph{KIN}\\
\end{cases}.
\end{equation}

Note, a $\mathbb{P}_s$ consists of a pair of templates and, thus, the task is to decide whether the media of the templates provide evidence that the two persons are blood relatives; notice Eq.~(\ref{eq:indicator}) is the template \emph{matcher} defined in Eq.~(\ref{eg:matcher}).

\newcolumntype{g}{>{\columncolor{Gray}}c}
\begin{table}[!t]
      \centering
        \scriptsize
        \caption{{Identification results for \emph{Image}, \emph{Image + Video}, and \emph{Image + Video + Audio}. Specifically, accuracy as a function of rank and mAP scores. The higher, the better.}}\label{subtab:task3:results}


	\begin{tabular}{m{14mm}rcccccc} 
	         &&  \multicolumn{5}{c}{\textbf{Rank}{\cellcolor{gray!15}}} &\tabularnewline
	\textbf{Modality}	&\textbf{Fusion}	& \textbf{@1} &	 \textbf{@5} & \textbf{@10}	&  \textbf{@20} &  \textbf{@50} &  \textbf{mAP}\tabularnewline

\rowcolor{green!10}\emph{\textbf{Image}} &\textbf{Mean} & 0.29 & 0.43 & 0.54 & 0.64 & 0.78 & 0.13\tabularnewline
   \rowcolor{green!10}  & \textbf{Median} & 0.28 & 0.44 & 0.52 & 0.64 & 0.77 & 0.13\tabularnewline
 \rowcolor{green!10}    & \textbf{Max} & 0.11 & 0.19 & 0.28 & 0.34 & 0.52 & 0.06\tabularnewline
   \rowcolor{green!10}  	 & 
   \textbf{TA} & 
   0.31 &
   0.43 &
   0.52 &
   0.63 &
   0.74 &
   0.14\tabularnewline

   \emph{\textbf{Image}} & \textbf{Mean} & 0.30 & 0.44 & 0.52 & 0.64 & 0.77 & 0.14\tabularnewline
   \emph{\textbf{+ Video}}  & \textbf{Median} & 0.28 & 0.44 & 0.50 & 0.63 & 0.76 & 0.14\tabularnewline
     & \textbf{Max} & 0.13 & 0.21 & 0.26 & 0.30 & 0.44 & 0.06\tabularnewline
          & 
          \textbf{TA} & 
          0.34 & 
          0.46 & 
          0.55 & 
          0.68 & 
          0.75 & 
          0.16\tabularnewline
     
     \rowcolor{green!10}   \emph{\textbf{Image}} & \textbf{Mean} & 0.30 & 0.44 & 0.52 & 0.64 & 0.77 & 0.14\tabularnewline
     \rowcolor{green!10}\emph{\textbf{+ Video}}  & \textbf{Median} & 0.28 & 0.44 & 0.50 & 0.63 & 0.76 & 0.14\tabularnewline
     \rowcolor{green!10} \emph{\textbf{+ Audio}} & \textbf{Max} & 0.13 & 0.21 & 0.26 & 0.30 & 0.44 & 0.06\tabularnewline
    \rowcolor{green!10}  &
     \textbf{TA} & 
     \textbf{0.56} &
     \textbf{0.59} &
     \textbf{0.63} &
     \textbf{0.74} &
     \textbf{0.78} & 
     \textbf{0.24}
\end{tabular} 
\end{table}

The data can be further split by the type of relationship $l$-- $L$ sets of pairs organized into $L$ independent lists, meaning $\mathbb{P}_s$ partitioned into $l$ disjoint sets in $\mathbb{P}_s^l$ (Table~\ref{tbl:counts}). Specifically, pair-types are \ac{bb}, \ac{ss}, or \ac{sibs} of mixed-sex; \ac{fd}, \ac{fs}, \ac{md}, or \ac{ms}; \ac{gfgd}, \ac{gfgs}, \ac{gmgd}, or \ac{gmgs}; four \emph{great grandparent} types as well. Hence, $l\in L\rightarrow\{BB, SS, \dots, GMGD, GMGS\}$ with $L=11$. Statistics for all relationship types are listed in Table~\ref{tab:data-counts}, with the lists of \emph{grandparent}-\emph{grandchild} and \emph{great-grandparent} omitted from \dif{the} verification task due to insufficient sample counts (\ie $L=7$).

As described, \ac{fiwmm} is organized as templates with many samples from various modalities (\ie still-face, face \dif{tracks,} audio, and transcripts (contextual)). Specifically, true IDs $y$ are paired with a template of all media available for the respective subject. In contrast with conventional kinship recognition, where one image is compared to another, the 1:1 paradigm is template-based (\ie one template compared to another). Then, a template pair $\mathbb{P}_s^l=((X_i, y_i), (X_j, y_j))$ is of different subjects (\ie $X_i$ and $X_j$, where $i\neq j$).

\subsubsection{Metrics}
\Ac{det} \diffy{curves and the average verification accuracy measured performances in verification, along with the \ac{tar} scores at specific values of \ac{far}} (Table~\ref{subtab:task1:results}).

\Ac{det} curves show error rates for binary classification systems, plotting the \ac{fnr} as a function of \ac{far}.
Type II error \ac{fnr} contributes to the Type I error \Ac{fpr}, revealing the \emph{imposter} accepted: 
$$
\text{FAR} =  \text{FPR}=\frac{\text{FP}}{{N^-}}=\frac{{FP}}{\text{FP}+\text{TN}},
$$
where the number of negatives is $N^-$. The geometric relationships of the metrics related to the score distributions and the choice in threshold show the trade-offs in error rates (\ie Type I versus Type II error):

$$
\text{FNR} =\frac{\text{FN}}{{N^+}}=\frac{{FN}}{\text{FN}+\text{TP}}.
$$
\diffy{Finally, $\ac{tar} + \ac{fnr}=1$; therefore, $\ac{tar}= 1 - \ac{fnr}$.}

In summary, various attempts were made for both \emph{genuine} and \emph{imposter} pairs, and the scores were saved. Then, by changing the threshold that converts the score to a decision, we could visualize the trade-offs between the different error types (\ie a lower threshold means fewer rejections of \emph{genuine} \dif{pairs} but more accepting of \emph{imposter} pairs). Thus, the performance of the system is highly dependent on a choice in \dif{the} threshold, which is the reason DET curves are used in biometrics to see these trade-offs in binary problems.

\begin{figure}[t!]
\centering
  \includegraphics[width=\linewidth]{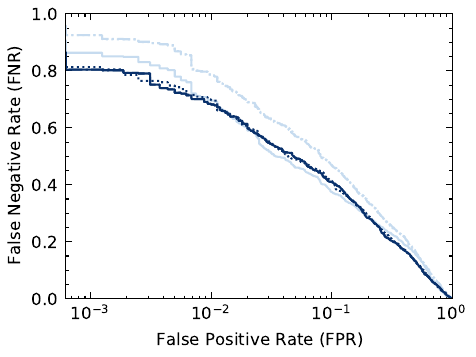}\\
  \includegraphics[width=\linewidth]{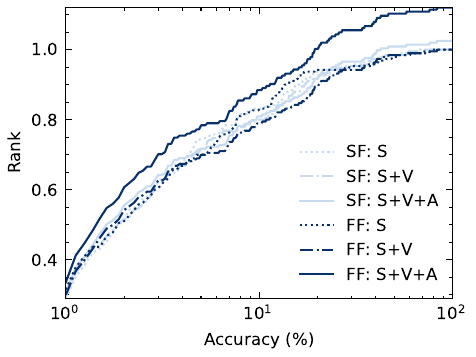}
  \caption{\textbf{Results for fusing early (\ie feature fusion (FF)) and late (\ie score fusion (SF)).} Still-images $S\text{,}$ video clips $V\text{,}$ and audio segments $A\text{,}$ with images and video $S+V$ and images, audio, and video $S+V+A\text{.}$ Both tasks gain from early fusion: the DET curve (top) summarizes the verification task by plotting FNR versus FPR (\ie lower is better); search \& retrieval shows a CMC curve (bottom)-- accuracy across ranks (\ie higher is better).}\label{fig:plots}
\end{figure}

\subsection{Search \& retrieval (missing family member)}
Kinship identification is organized as a 1:N search and retrieval task, with each subject having one-to-many media samples. Thus, we imitate template-based evaluation protocols~\cite{maze2018iarpa}. Furthermore, the goal is to find relatives of search subjects (\ie \textit{probes}) in a search pool (\ie \textit{gallery}).

\subsubsection{Data splits and settings}
A gallery $G=\{g_i\}, ( i=1,...,N)$ is queried by a set of probes $P=\{p_j\}, ( j=1,...,M)$ for search and retrieval, where $g_i$ is the $i^{th}$ template in $G$ and $p_j$ is the template of the $j^{th}$ query subject. As mentioned, a template consists of samples of various modalities. Given a template of \ac{mm}, various schemes were applied to integrate the identity information from all media components of $P\text{.}$

\subsubsection{Metrics} 
Scores of $N$ missing children are calculated as
$$
AP(l)=\frac{1}{P_L}\sum^{P_L}_{tp=1}Prec(tp)=\frac{1}{P_L}\sum^{P_L}_{tp=1}\frac{tp}{rank(tp)},
$$
where \ac{ap} is a function of family $l\in L$ (\ie $|L| = {P_L}$) for a given \ac{tpr}. All \ac{ap} scores are averaged to find the mean \ac{ap} (\ie mAP):
$$mAP = \frac{1}{N}\sum^{N}_{l=1}AP(l).$$

Also, \ac{cmc} curves as a function of rank enable for analysis between different attempts~\cite{decann2013relating}, along with the accuracy at rank 1, 5, and 10.

Our choice in metrics is typical for judging \dif{the} ranking capabilities of classification (or identification) systems. Like the DET assesses the 1:1 case of verification, CMC measures the 1:$m$ performance in ranking-based problems.

\section{Benchmarks}
\subsection{Methodology}

The problems of \ac{fiwmm} have various views– multi-source and \dif{multimodal.} The former varies in samples and \dif{treats} the different \dif{media types} independently until the matching function outputs scores (\ie late-fusion). The latter demands a method for \dif{the} early fusion \dif{(\eg feature-level), enhancing} performance by using informative samples while ignoring noisy and less discriminating samples. We next describe the modality-specific features \dif{(\ie encoding different media types)} and methods of fusion.

\subsubsection{Visual features}
\dif{We represented each visual media sample} as the encoding from Arcface CNN~\cite{deng2019arcface} (\ie ResNet-34). MS1M~\cite{guo2016ms} was the train set, which had $\approx$5.8M faces for 85,000 subjects. Faces were detected with \dif{MTCNN~\cite{zhang2016joint},} returning the five facial landmarks (\ie two eyes, nose, both corners of the mouth). Then, faces were cropped and aligned using a similarity transformation on the five points with references set by the eye locations. Once \dif{cropped, we resized the faces} to 96$\times$112. The RGB (\ie pixel values of [0, 255]) were center about 0 (\ie subtracting 127.5) and then standardized  (\ie divided by 128). \dif{We passed images through a CNN to map to features--} all \dif{features} were later L2 normalized~\cite{wang2017normface}. During training, the  \dif{batch size}  was  200, and \dif{an}  SGD optimizer with a momentum \dif{of} 0.9, weight decay 5e-4, and learning rate starting at 0.1 and decreasing 10$\times$ twice, both times when the error leveled. \dif{We followed the settings of} the \ac{sota} Arcface-- a popular choice for an off-the-shelf choice for \ac{fr} technology and applications.\footnote{Followed \href{https://github.com/deepinsight/insightface}{https://github.com/deepinsight/insightface}.}

\dif{We processed} both images and videos as described, with the mapping from image-space to feature-space denoted by $\mathcal{F}(x)\in\mathbb{R}^\mathrm{d}$, where the dimension $\mathrm{d}=512$ for ArcFace. \dif{Face tracks from video data are fused to a single encoding by average pooling the same track features. We represent} a face track of $M$ frames as $\bar{z}=\frac{1}{M}\sum_x\mathcal{F}(x)$. This reduces the effect of noisy \dif{frames} and smooths out the final representation to weigh \dif{like} all other media samples.

\subsubsection{Audio features} 
\dif{We encoded all speech segments} with a \ac{sota} deep learning \dif{model~\cite{chung2020in}.} Specifically, we trained  SqueezeNet~\cite{iandola2016squeezenet}, a 34-layer ResNet~\cite{he2016deep}, with an \emph{angular prototypical loss} and optimized with Adam~\cite{kingma2014adam} to transform WAV-encoded audio files to a single encoding (\ie $f(x)=z\in\mathcal{R}^{512}$). \dif{Thus, per the} \emph{angular prototypical loss}~\cite{snell2017prototypical}, used alongside softmax, minimizes within-class scatter (\ie penalty formed as the sum of \dif{Euclidean} distances from all samples of a subject from the mean centroid of the respective mini-batch). \dif{Expressly,} a support set $S$ and a query $Q$ are set in each mini-batch on a subject-by-subject basis, with $Q$ \dif{made up} of a single utterance to compare with the centroid of $S$ that consists of all other samples in the mini-batch for that class. \emph{Angular prototypical} takes advantage of the perks of using centroid \dif{prototypes} while enhancing by following \ac{ge2e}~\cite{wan2018generalized} usage of a cosine-based similarity metric. \dif{This is scale-invariant, is more robust to feature variance, and helps convergence during training~\cite{wang2017deep}.}

\diff{\subsubsection{Naive fusion}\label{subsec:fusion}
We show results from \dif{various} naive fusion techniques (\eg average pooling of features and voting of scores). To no surprise, the score-based fusion outperforms the feature-level fusion schemes. \diffy{Specifically, the mean of all scores, both within a template and comparing templates, further improved by adding the other media types (Table~\ref{subtab:task1:results}).} The gain from each added modality is clear from just the naive score-fusion. Still, the naive fusion methods at the feature level are an ineffective way of combining knowledge. Provided media - media that \dif{vary} in  modality, \dif{quality,} and discriminative power - a simple, unweighted average across the items of a template does not exploit all information. To better fuse the template, we adapt a model to the template to better discriminate family members.}

\subsubsection{Feature fusion}\label{subsec:featfusion}

\Ac{ta}~\cite{whitelam2017iarpa} is a form of transfer learning that fuses labeled face \dif{features} from a source domain with \dif{template-specific} \acp{svm} trained on the target domain. \dif{We followed \emph{probe adaptation} settings for kinship verification;} for identification (\ie search \& retrieval), we follow \emph{gallery adaptation} settings. Regardless of the setting, the goal is to train a similarity function from a probe template $P$ to either a reference template $Q$ (\ie similarity $s(P,Q)$) or gallery $G$ (\ie similarity $s(P,G)$) for \emph{probe adaptation} and \emph{gallery adaptation}, respectively. Thus, given $P$, we train an \ac{svm} on top of its encoded media in \dif{$x^+$ and $x^-$ as the positive and negative, respectively:} for \emph{probe adaptation}), the negative set is formed by sampling a single sample from all other templates to use for $P$ and $Q$; for \emph{gallery adaptation}, the negatives are of templates in $G=\{X_1,X_2,\dots,X_m\}$ besides the $i^{th}$ template $X_i\neq X$. Either way, \ac{ta} allows for the early fusion of different media types with many more negative samples (\ie $x^+ \ll x^-$). The number of positive samples is minimal (\ie $|P|<<|N|$); hence, the \ac{svm}-based modeling mitigates imbalanced data about the number of positives versus negatives. \dif{Notably, a kernel space defined by the largest margins found by \acp{svm} handled cases with one or few samples the best.}

Finally, let $P(q)$ represent the evaluation of media \dif{features} of template $Q$ upon being trained on $P$, and vice versa for $Q(p)$ to use knowledge from both \dif{directions:}
\begin{equation}
s(P,Q) = \frac{1}{2}P(q) + \frac{1}{2}Q(p).
\end{equation}
The similarity score is the result of the templates fused.

The benefit of \acp{svm} is in the kernel. Specifically, the linear, max-margin modeling scheme of a vanilla \ac{svm} has proven effective at separating a non-linear feature space between two classes; (\ie $i$ and $j$, where $y_{ij}=\pm1$ for the same ($+$) and different ($-$) classes). Thus, the implicit embedding function (\ie kernel) $K(x_i, x_j, y_{ij})=\varphi(x_i, y_i)\varphi(x_j, y_j)$ projects the encoding pair to a non-linear space such that the \ac{svm} learns the best hyperplane $\mathbf{w}^T K(x_i, x_j, y_{ij}) + b = 0$ separating the two classes. This is done on \dif{the} training set by (1) maximizing the margin and (2) minimizing the loss-- weights $\mathbf{w}$ are learned, while bias terms $b$ are set to 1 (\ie concatenated on $\mathbf{w}$, as an added dimension). Also, $K(x_i, x_j, y_{ij})=\exp{\frac{||x_i- x_j||^2}{2\sigma^2}}$ for ${y_{ij}\in\{-, +\}}$ as the respective class (\ie Gaussian RBF kernel~\cite{scholkopf2002learning} projects all \dif{features} to a higher dimension). Then, the predicted class is inferred as $\hat{y}=\mathbf{w}^T\varphi(x_i)\varphi(x_j) + b$. We used \emph{dlib's}~\cite{king2009dlib}-- L2 regularized cosine-loss with class-weighted hinge-loss, \ie
\newcommand\mycommfont[1]{\small\ttfamily\textcolor{blue}{#1}}
\SetCommentSty{mycommfont}

\begin{figure}[!t]
\centering
    \includegraphics[width=\linewidth]{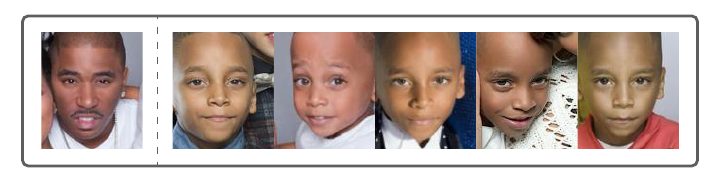}
  \caption{Template of \dif{an actual} FS pair incorrectly classified using score fusion, but correct for TA (\ie feature fusion). Only a single face is available for the father (left), while all \dif{son instances} are at an early age (right).}
  \label{fig:fs:visual:qual}
\end{figure}

  \begin{align}
    \nonumber \min_\mathbf{w}{\frac{1}{2}\mathbf{w}^T\mathbf{w}} + \lambda_{+}\sum_{i=1}^{N_{+}}\max\large{[}0, 1 - y_i\mathbf{w}^Tf(x_i)\large{]}^2 \\
    + \lambda_{-}\sum_{j=1}^{N_{-}}\max\large{[}0, 1 - y_j\mathbf{w}^Tf(x_j)\large{]}^2.
    \label{eq:hinge}
 \end{align}

The protocols are set for \emph{gallery adaptation}: train a similarity function $s(P, G)$ from a probe $P$ to gallery $G$. A gallery of templates $G=\{X_1, X_2, \dots, X_m\}$ are used to train the \ac{svm} (\ie the scoring function $s(P, X_i)$). The difference between \emph{probe adaptation} and \emph{gallery adaptation} is in the negative sets. Along with the sample per subject trained against for \emph{probe adaptation}, \emph{global adaptation} samples all other templates in $G$ as added negatives. Again, $N_{+} << N_{-}$. The class imbalance is handled via class-weighted hinge-loss in Eq~\ref{eq:hinge}, with $\lambda_{+}=\lambda\frac{N_{+} + N_{-}}{2N_{+}}$, $\lambda_{-}=\lambda\frac{N_{+} + N_{-}}{2N_{-}}$, which are regularization constants  inversely proportionate to class frequency. The constant $\lambda$ trades-off between the regularization and loss, which we set to 10 as in earlier work~\cite{whitelam2017iarpa}.

\subsubsection{Implementation}
The system was implemented in Python, with CNNs for visual and audio from \dif{PyTorch's} deep learning framework for each media encoder and \ac{svm} from LibSVM. The negative sample set was formed by randomly selecting a single instance from all other families. Hence, investigating and improving the naive means for which we pool negative samples is a promising direction for future work. With that, $N_{+} \text{ and } N_{-}$ \dif{are} set case-by-case.

\subsection{Results}
System performance boosted with each added modality (Fig.~\ref{fig:plots}, Table~\ref{subtab:task1:results} and~\ref{subtab:task3:results}). Considering the benchmarks use conventional speech and \ac{fr} \dif{technology} and our hypothesis that video and audio \dif{boost} discrimination, these notable improvements would likely continue to climb supplied a more sophisticated or specific solution. It would be interesting to fuse earlier \dif{on} and train machinery jointly for \dif{audiovisual}  data. From this, more complex dynamics of facial appearance, along with the corresponding speech signal, could further our knowledge and \dif{supply} insights.

\begin{figure}[!t]
\centering
    \includegraphics[width=\linewidth]{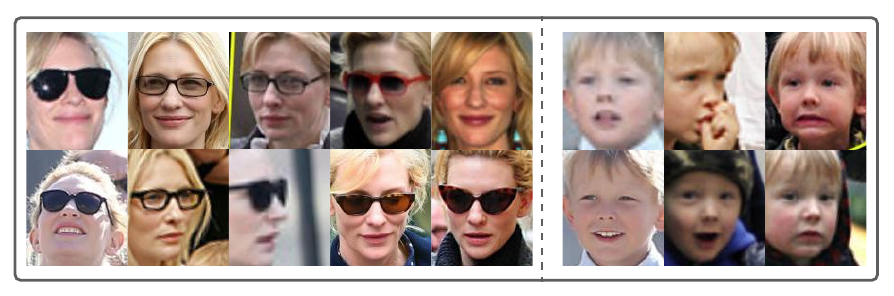}
  \caption{A MS pair incorrectly classified with late fusion but correct with early fusion (\ie TA). The young age of the son and most faces of the mother occluded by sunglasses pose a challenge (\ie score fusion puts equal weight on all samples, where TA learns \dif{to discriminate better).}}
  \label{fig:ta:visual:qual}
\end{figure}

There was a trend in the type of corrected samples when comparing the score-based fusion to the feature-based (\ie \ac{ta}). As shown in Figure~\ref{fig:fs:visual:qual}, the challenge of recognizing kinship from samples of one or more \dif{members} at an early age is mitigated. \ac{ta} learns to better discriminate in these conventional failure cases. Additionally, some templates with multiple instances are often better than others when comparing. Hence, \ac{ta} does not simply average all instances equally, like for naive score-based fusion-- fix cases with few samples are most discriminate (Fig.~\ref{fig:ta:visual:qual}).

\subsection{Discussion}
The template-based protocol adds practical value by mimicking the more likely structure posed in operational settings, per \ac{nist}~\cite{maze2018iarpa}. Besides, several other factors make it a more exciting formulation \dif{and,} therefore, a higher potential for researchers to \dif{show off} their creativity. For instance, opposed to using a single sample per subject (\ie one-shot learning), each now is represented in a set of media (\ie a template). The question now \dif{arises} - how to best fuse knowledge and incorporate evidence from different modalities and how to best learn from all available MM data? Another consequence of using templates is that the random chance is increased from (1) the knowledge added to \dif{the} pool (or fuse) from the added modalities, and (2) the gallery size reduces from tens of thousands by nearly ten-fold. The latter is not an implication of lessor \dif{difficulty} but the byproduct of reducing bias in data~\cite{robinson2020face}. That is, opposed to having one-to-many samples per \dif{subject,} there is just one template. Mitigating specific sources of data imbalance (\ie whether there are thirty samples or just \dif{one),} a system’s ability to recognize a particular pairing or group affects the metric evenly. In other words, a system may easily recognize a specific parent-child pair - regardless of the number of face samples and, so, the number of face pairs. Hence, the impact on the metric is proportional to the number of unique pairs. 

Furthermore, for verification, we measured the impact of the different results using the \dif{significant} test. Expressly, we set the baseline image-only as the null hypothesis to compare an alternative hypothesis set as the different results (\ie image-only, image and video, and \dif{audiovisual).} Specifically, comparing baseline to null results in a p-value of 1.000 (\ie same results). Then, a limited improvement p-value=0.974 (\ie \dif{the smaller, the} better) results compared to the mean of image and videos (\ie videos add samples, but of the similar modality after averaging frames). Significant improvement in p-value=0.024 \dif{compared} with the naive fusion of MM (\ie audio and visual), which goes to 0.000 for TA using MM. Thus, this further confirms our claim that kinship recognition systems significantly better the decisions.

\section{Future Work}
\ac{fiwmm} pushes the bar for possibilities in automatic recognition of kinship in \ac{mm}. An immediate next step for research involves \dif{gathering} experts of different domains, such as those in sequence-to-sequence modeling, whether video (\ie visual), audio (\ie speech), contextual (\eg conversations and parts-of-speech), or early fusing pairs or groups. Let us next discuss a variety of ways that we foresee the data being beneficial and to form \dif{a} collaboration amongst different research \dif{communities} and beyond (\ie \diff{although FIW-MM is for non-commercial uses, the resource has high} commercial potential and, \diff{thus, \emph{proof of principle} experiments can motivate business moves).}

We expect \ac{fiwmm} to bring anthropology and genealogy experts together with researchers of \ac{mm} and ML to help find the hidden patterns connecting families in \ac{mm}. For instance, we showed that \dif{simply} applying pre-trained models from the speech recognition domain allows for the audio signal to be incorporated for more discriminative power than visual evidence alone. Indeed, there is room to improve over the simple benchmark. Furthermore, high-level semantics (\ie attributes) like accents, commonly used phrases, and speaker demeanor could boost the overall performance and supply insights by interpretation. Similarly, studies on familial language components and inherited changes, or even deltas across the same generation (\ie commonalities and differences in siblings’ speech), can too be quite revealing. A similar potential exists for the videos and audiovisual data in model complexity and use-cases.

The family trees, \dif{an} abundance of data points, rich metadata for \dif{individuals,} and relationships among \ac{mm} data -- \ac{fiwmm} could serve as a basis for group-based (\ie social) data mining. More data can enhance or target specific nature-based studies, traditional ML-based audio, visual, and \dif{audiovisual}  tasks, or even extend this dataset. Fusing \dif{audiovisual}  data is an ongoing, unanswered problem~\cite{song2019review}. Note that FIW-MM in its entirety pose more problems than it solves: from the model \dif{training} to improvements made when dealing with missing or incomplete modalities, and even the data processing and data imbalance; from the underlying roots of the problem to the high-level semantics, similar to modern biometrics systems with \dif{audiovisual}  data, \ac{fiwmm} is an appropriate step considering the state in visual kinship recognition technology.

Another direction is fusion. We included early and late fusion by merging the media as features and scores, respectively. Scores were fused by naively averaging-- ignoring the signal \dif{type} and assuming all samples should carry the same weight. Fusion can incorporate more sophisticated techniques: cross-modality, selectively choosing the best quality \dif{samples or model-based} decision trees. This concept, alone, is vastly in need of solutions-- whether data fusion, where the input is clips of aligned \dif{audiovisual;} early-fusion, as we did via \ac{ta} to fuse at the feature-level; or late-fusion, which we also included by naively averaging scores. Besides, meta-knowledge, like relationship types (\eg directional relationships), genders, age, and other attributes, could indicate final decisions. Hence, there \dif{is} an abundance of fusion paradigms-- none are trivial, yet most hold promise. 

\begin{figure}[t!]
    \centering
    \includegraphics[width=.88\linewidth]{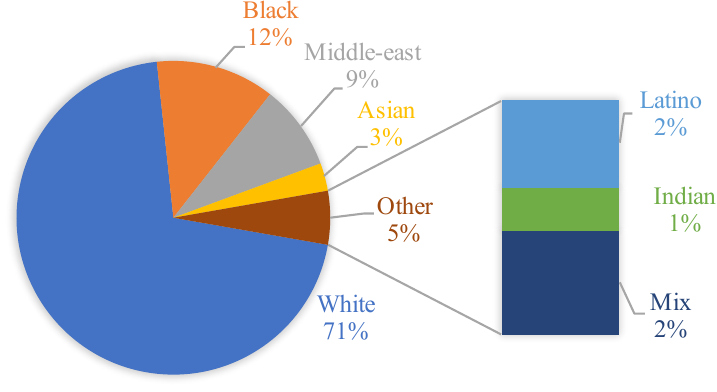}
    \caption{Ethnicity distrubution of the families in \ac{fiw}. This is based on the output of pretrained ethnicity classify and,  hence,  is only an approximation.  We used a public github, \href{https://github.com/shaoanlu/face_toolbox_keras}{face\_toolbox\_keras},  to acquire predicted labels.}\label{fig:ethnicities}
\end{figure}

Besides, questions concerning bias-- trends as a function of gender and ethnicity; properly securing family data, addressing areas of privacy and protection. Furthermore, studies on differences between a diverse pool versus a search pool of mostly similar faces-- whether it be a closer look at the effects of age for different relationship types, quantifying the similarity of specific features across different subgroups, and its effects on appearance (\ie visual data) and speech (\ie audio data).  As has been a focus in recent works~\cite{robinson2021balancing, DBLP:journals/corr/abs-2102-02320}, considerations of bias across subgroups,  along with attempts to acquire a balanced set of families with respect to individual demographics~\label{fig:ethnicities}.  

Research topics to spawn off the proposed are vast; the specifics suggested here are limited by our imagination. We expect scholars and experts of other domains to see paradigms not mentioned here: whether it be an improved variant of adapting templates and feature fusion (\eg~\cite{xiong2017good}), deciding when to fuse, a new method of integration, along with the integration details, are all open research questions.

\section{Conclusion}
We extended \dif{the} \acf{fiw} image dataset for kinship recognition with multimedia (MM) data-- 1-3 MM samples for 2-4 members from 200 (of 1,000) families we \dif{obtained} and then renamed \acf{fiwmm}, which is the first dataset to provide multimedia data for families in ML-related fields. In addition, \dif{we followed} new paradigms (\ie template-based protocols) in both benchmarks (\ie kinship verification and search \& retrieval of family members)-- templates mimic a realistic setting, as followed in \ac{fr}-related problems, but with this a first for kin-based recognition. Our labeling pipeline uses \dif{multimodal} evidence and a simple feedback schema to use the labeled data of \ac{fiw} to propagate ground truth for the added modalities. \diff{Results improve with each added media type, with the top performance obtained with \dif{an} early fusion of features of multiple modalities. \ac{fiwmm} marks a \dif{significant} milestone for kin-based problems by welcoming experts of other data domains.} In addition, \ac{fiwmm} supports \dif{several} recognition tasks due to its rich metadata, template-based \dif{structure,} and multiple modalities.

{
\bibliographystyle{IEEEtran}
\bibliography{IEEEabrv,sample-base}
}

\vspace{-15mm}
\begin{IEEEbiography}[{\includegraphics[width=1in,height=1.25in,clip,keepaspectratio]{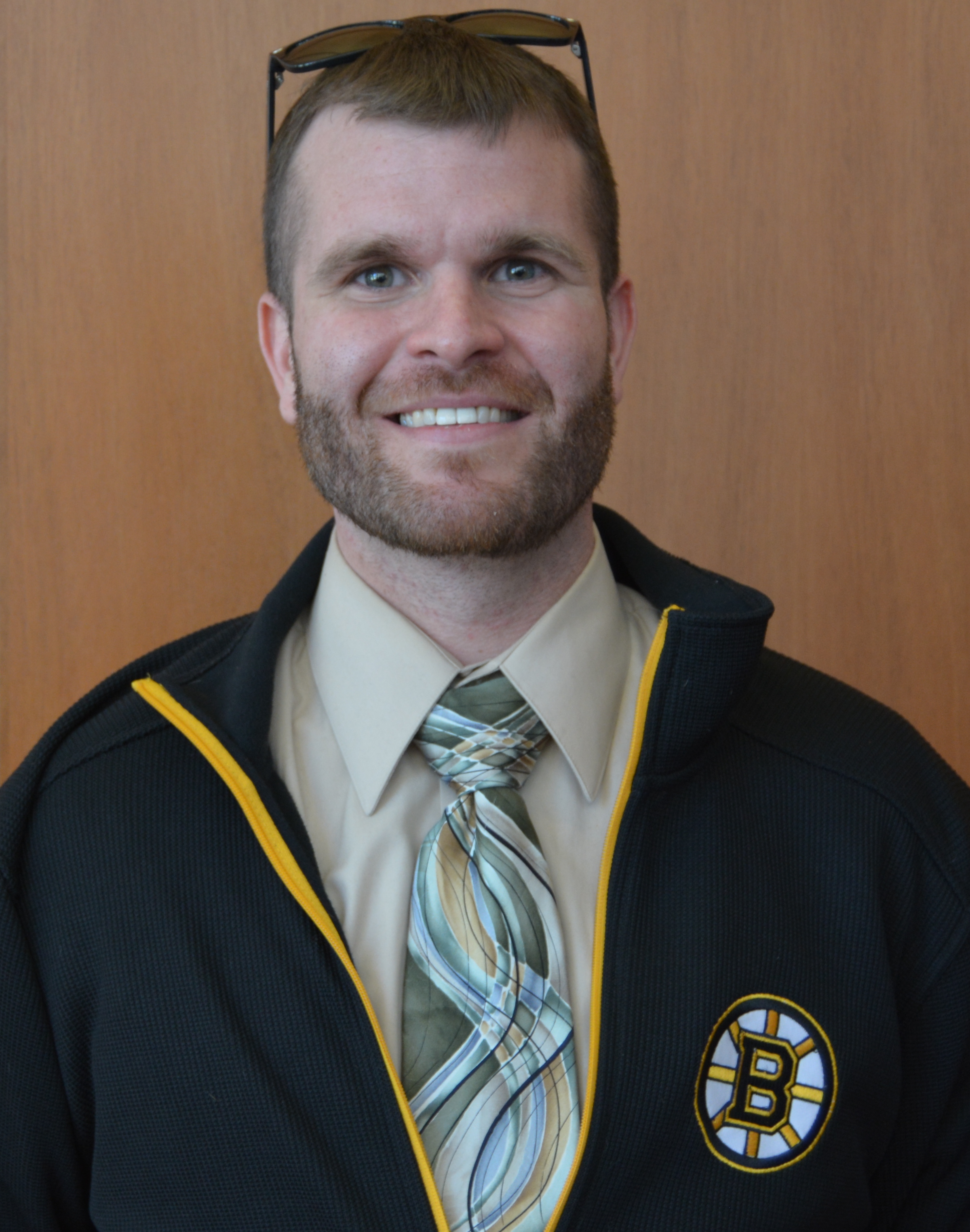}}]{Joseph P Robinson}
B.S. in electrical \& computer engineering ('14) and Ph.D. in computer engineering ('20) at Northeastern University, where he worked as part-time faculty: taught undergrads in Data Analytics ('19-'20~\emph{Best Teacher}). Research is in applied machine vision, emphasizing faces, deep learning, MM, and big data. He led on TRECVid debut (MED'15, third place). Built many images and video datasets-- most notably FIW. Organized \& hosted several workshops and challenges (\eg NECV17, RFIW@ACMMM17, RFIW@FG18-20, AMFG@CVPR18, FacesMM @ICME18-19), tutorials (ACM-MM18, CVPR19, FG19), PC member (\eg CVPR, FG, MIRP, MMEDIA, AAAI, ICCV, \etc), reviewer (\eg IEEE TBioCAS, TIP, TPAMI,\etc), and Pres. of IEEE@NEU and Rel. Officer of IEEE SAC R1 Region. Completed NSF REUs ('10 \& '11); co-op at Analogic Corp. ('12) BBN Tech. ('13); intern at MIT LL ('14), STR ('16-'17), Snap Inc. ('18), ISMConnect ('19). Currently, an AI Engineer at Vicarious Surgical ('21), working on surgical robotics.
\end{IEEEbiography}
\vspace{-12mm}

\begin{IEEEbiography}[{\includegraphics[width=1in,height=1in,clip,keepaspectratio]{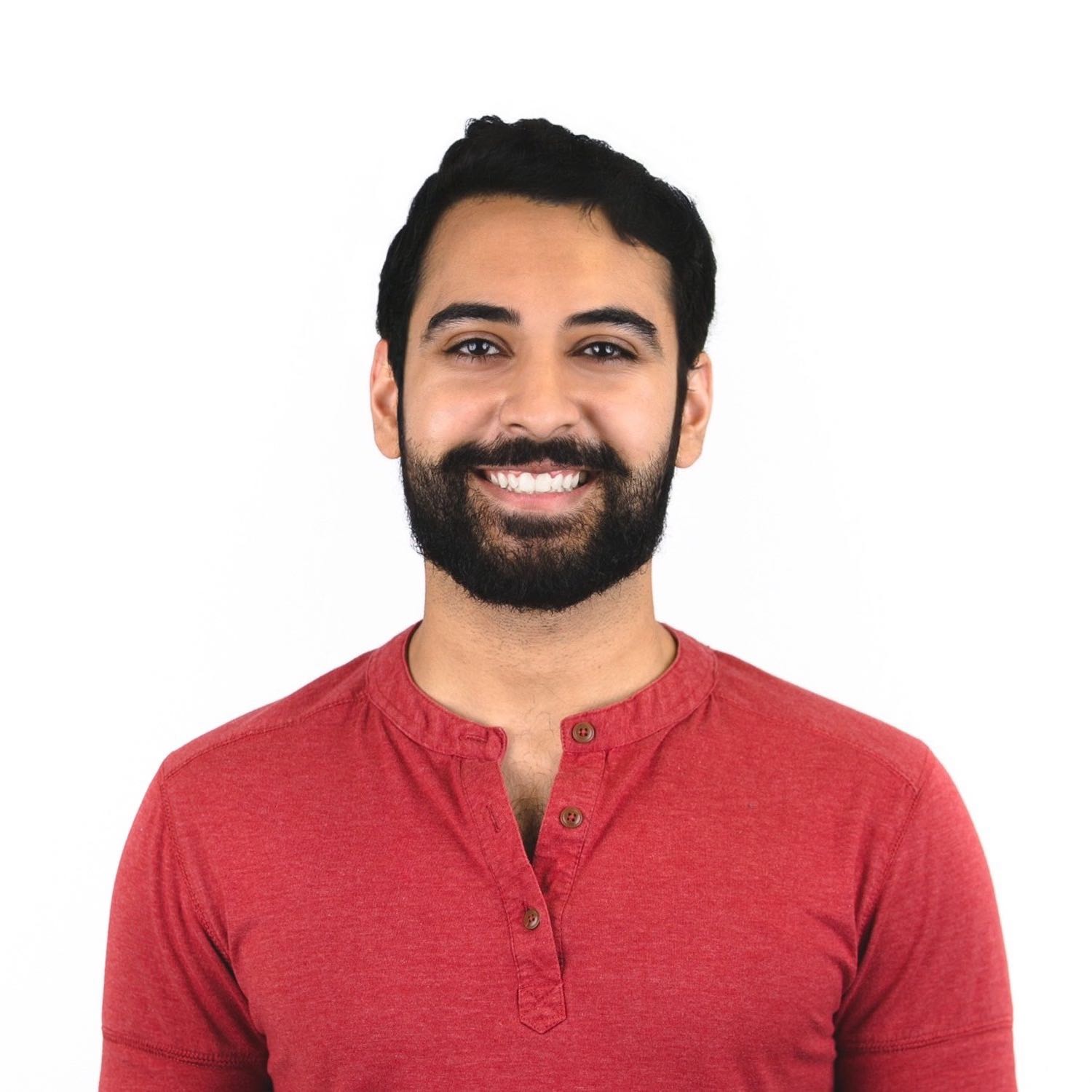}}]{Zaid Khan}
Received a  B.S. in Computer engineering ('18) and is currently a Ph.D. student in Computer Engineering ('25) at Northeastern University under Dr. Yun Fu. Zaid served as the website and technical chair of RFIW 2020 held in conjunction with IEEE Conference on Facial and Gesture Recognition. His research interests lie in computer vision, with a focus on faces. 
\end{IEEEbiography}

\vspace{-12mm}

\begin{IEEEbiography}[{\includegraphics[width=1in,height=1.25in,clip,keepaspectratio]{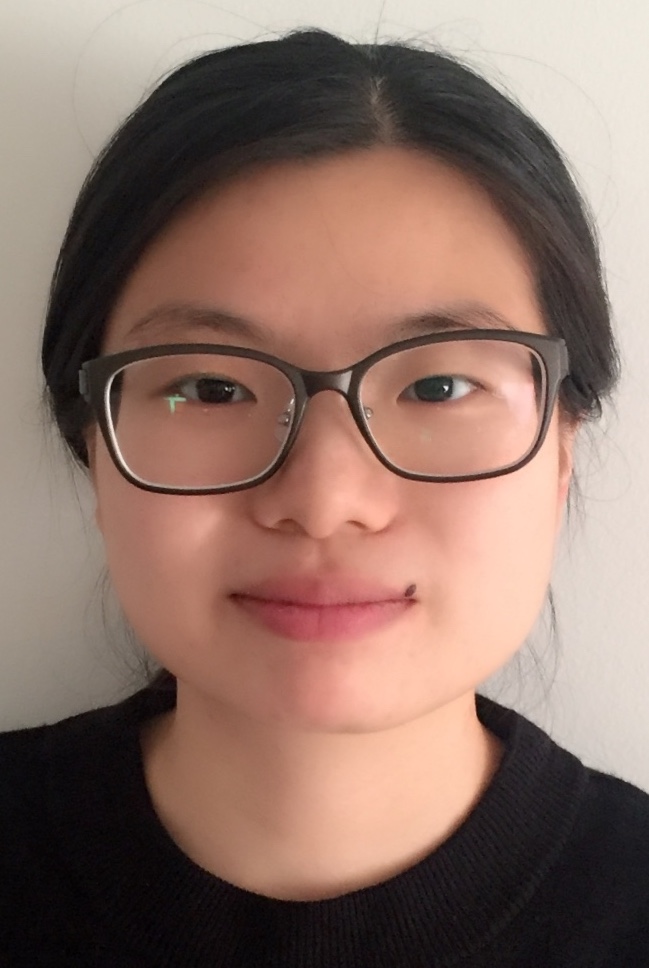}}]{Yu Yin}
Received B.S. in Electrical and Information Engineering (2016) from Wuhan University of Technology, China, M.S. in Electrical and Computer Engineering (2018) at Northeastern University (NEU), Boston, MA, and is currently pursuing a Ph.D. in Computer Engineering at NEU under Dr. Yun Fu. Her research spans image processing (\ie super-resolution, face generation), visual recognition (\ie face recognition, pose estimation, face alignment, and emotion recognition), and biosignal processing. She served as organizing chair of RFIW Workshop Challenge @ FG'20, conference PC (\eg AAAI \& FG), and journal reviewer (\eg IEEE Trans. on TIP \& TNNLS). She interned at Zillow Group (2020). 
\end{IEEEbiography}

\vspace{-12mm}

\begin{IEEEbiography}[{\includegraphics[width=1in,height=1.25in,clip,keepaspectratio]{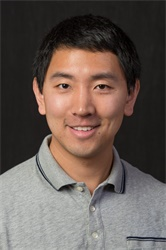}}]{Ming Shao}
Received the B.E. degree in computer science, the B.S. degree in applied mathematics, and the M.E. degree in computer science from Beihang University, Beijing, China, in 2006, 2007, and 2010, respectively. He received the Ph.D. degree in computer engineering from Northeastern University, Boston MA, 2016. He is a tenure-track Assistant Professor affiliated with College of Engineering at the University of Massachusetts Dartmouth since 2016 Fall. His current research interests include sparse modeling, low-rank matrix analysis, deep learning, and applied machine learning on social media analytics. He was the recipient of the Presidential Fellowship of State University of New York at Buffalo from 2010 to 2012, and the best paper award winner/candidate of IEEE ICDM 2011 Workshop on Large Scale Visual Analytics, and ICME 2014. He has served as the reviewers for many IEEE Transactions journals including TPAMI, TKDE, TNNLS, TIP, and TMM. He has also served on the program committee for the conferences including AAAI, IJCAI, and FG. He is the Associate Editor of SPIE Journal of Electronic Imaging, and IEEE Computational Intelligence Magazine. He is a member of IEEE.
\end{IEEEbiography}

\begin{IEEEbiography}[{\includegraphics[width=1in,height=1.25in,clip,keepaspectratio]{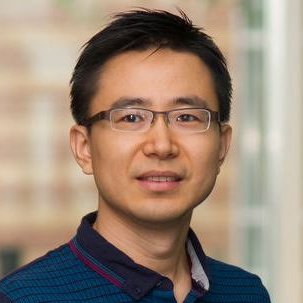}}]{Yun Fu}
(S’07-M’08-SM’11-F’19) received the B.Eng. degree in information engineering and the M.Eng. degree in pattern recognition and intelligence systems from Xi’an Jiaotong University, China, respectively, and the M.S. degree in statistics and the Ph.D. degree in electrical and computer engineering from the University of Illinois at Urbana-Champaign, respectively. He is an interdisciplinary faculty member affiliated with College of Engineering and the College of Computer and Information Science at Northeastern University since 2012. His research interests are Machine Learning, Computational Intelligence, Big Data Mining, Computer Vision, Pattern Recognition, and Cyber-Physical Systems. He has extensive publications in leading journals, books/book chapters and international conferences/workshops. He serves as associate editor, chairs, PC member and reviewer of many top journals and international conferences/workshops. He received seven Prestigious Young Investigator Awards from NAE, ONR, ARO, IEEE, INNS, UIUC, Grainger Foundation; eleven Best Paper Awards from IEEE, ACM, IAPR, SPIE, SIAM; many major Industrial Research Awards from Google, Samsung, Amazon, Konica Minolta, JP Morgan, Zebra, Adobe, and Mathworks, etc. He is currently an Associate Editor of the IEEE Transactions on Neural Networks and Leaning Systems. He is fellow of IEEE, IAPR, OSA and SPIE, a Lifetime Distinguished Member of ACM, Lifetime Member of AAAI, and Institute of Mathematical Statistics, member of Global Young Academy, INNS and Beckman Graduate Fellow during 2007-2008.
\end{IEEEbiography}

\end{document}